\documentclass[10pt,twocolumn,letterpaper]{article}

\usepackage{cvpr}
\usepackage{times}
\usepackage{epsfig}
\usepackage{graphicx}
\usepackage{amsmath}
\usepackage{amssymb}
\usepackage[ruled]{algorithm2e} 

\usepackage[pagebackref=true,breaklinks=true,letterpaper=true,colorlinks,bookmarks=false]{hyperref}

\cvprfinalcopy 

\def\cvprPaperID{11659} 

\begin{document}

\title{DeepTag: An Unsupervised Deep Learning Method for Motion Tracking on Cardiac Tagging Magnetic Resonance Images}


\author{Meng Ye$^{1}$, Mikael Kanski$^{2}$, Dong Yang$^{3}$, Qi Chang$^{1}$, \\
\vspace{12pt}
Zhennan Yan$^{4}$, Qiaoying Huang$^{1}$, Leon Axel$^{2}$, Dimitris Metaxas$^{1}$\\
$_{}^{1}\textrm{}$Rutgers University, $_{}^{2}\textrm{}$New York University School of Medicine, $_{}^{3}\textrm{}$NVIDIA, \\
$_{}^{4}\textrm{}$SenseBrain and Shanghai AI Laboratory and Centre for Perceptual and Interactive Intellgience\\

{\tt\small \{my389, qc58, qh55, dnm\}@cs.rutgers.edu }

}

\maketitle

\begin{abstract}
   Cardiac tagging magnetic resonance imaging (t-MRI) is the gold standard for regional myocardium deformation and cardiac strain estimation. However, this technique has not been widely used in clinical diagnosis, as a result of the difficulty of motion tracking encountered with t-MRI images. 
   In this paper, we propose a novel deep learning-based fully unsupervised method for in vivo motion tracking on t-MRI images. 
   We first estimate the motion field (INF) between any two consecutive t-MRI frames by a bi-directional generative diffeomorphic registration neural network.
   Using this result, we then estimate the Lagrangian motion field between the reference frame and any other frame through a differentiable composition layer.
   By utilizing temporal information to perform reasonable estimations on spatio-temporal motion fields, this novel method provides a useful solution for motion tracking and image registration in dynamic medical imaging.
   Our method has been validated on a representative clinical t-MRI dataset; the experimental results show that our method is superior to conventional motion tracking methods in terms of landmark tracking accuracy and inference efficiency. 
   Project page is at: \url{https://github.com/DeepTag/cardiac_tagging_motion_estimation}.
\end{abstract}

\section{Introduction}
\begin{figure}[t]
\begin{center}
\includegraphics[width=1.0\linewidth]{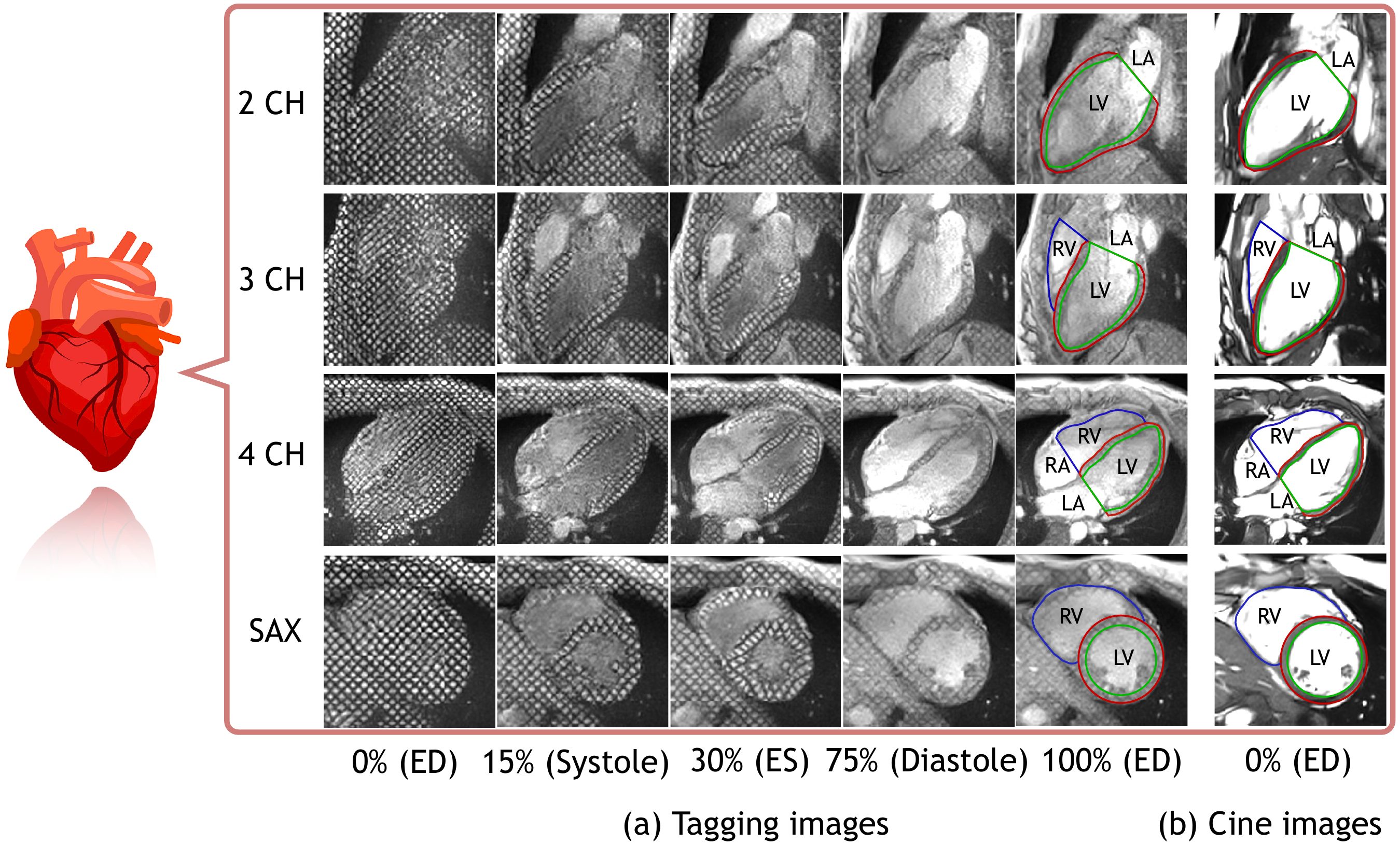}
\end{center}
   \caption{Standard scan views (2-, 3-, 4-chamber views and short-axis views) of cardiac MRI. (a) Tagging images. Number under the figure means percentage of one cardiac cycle. (b) End-diastole (ED) phase of cine images. Red and green contours depict the epi- and endo-cardial borders of left ventricle (LV) myocardium (MYO) wall. Blue contour depicts the right ventricle (RV). LA: left atrium. RA: right atrium.}
\label{fig1}
\end{figure}

Cardiac magnetic resonance imaging (MRI) provides a non-invasive way to evaluate the morphology and function of the heart from the imaging data. Specifically, dynamic cine imaging, which generates a 2D image sequence to cover a full cardiac cycle, can provide direct information of heart motion.
Due to the long imaging time and breath-holding requirements, the clinical cardiac MRI imaging protocols are still 2D sequences.
To recover the 3D motion field of the whole heart wall, typically we need to scan several slices in long axis (2-, 3-, 4-chamber) views and short-axis (SAX) views, as shown in Fig.~\ref{fig1}. There are two kinds of dynamic imaging: conventional (untagged) cine MR imaging and tagging imaging (t-MRI)~\cite{amzulescu2019myocardial}.
For untagged cine images (most recent work has focused on these images), feature tracking can be used to estimate myocardial motion~\cite{krebs2019learning,morales2019implementation,qin2018joint,zheng2019explainable,yu2020foal,yu2020motion}.
However, as shown in Fig.~\ref{fig1} (b), due to the relatively uniform signal in the myocardial wall and the lack of reliable identifiable landmarks, the estimated motion cannot be used as a reliable indicator for clinical diagnosis.
In contrast, t-MRI provides the gold standard imaging method for regional myocardial motion quantification and strain estimation.
The t-MRI data is produced by a specially designed magnetic preparation module called spatial modulation of magnetization (SPAMM)~\cite{axel1989mr}. It introduces the intrinsic tissue markers which are stripe-like darker tag patterns embedded in relatively brighter myocardium, as shown in Fig.~\ref{fig1} (a). By tracking the deformation of tags, we can retrieve a 2D displacement field in the imaging plane and recover magnetization, which non-invasively creates fiducial ``tags" within the heart wall.

Although it has been widely accepted as the gold standard imaging modality for regional myocardium motion quantification, t-MRI has largely remained only a research tool and has not been widely used in clinical practice. The principal challenge (detailed analysis in Supplementary Material) is the associated time-consuming post-processing, which could be principally attributed to the following: (1) Image appearance changes greatly over a cardiac cycle and tag signal fades on the later frames, as shown in Fig.~\ref{fig1} (a). (2) Motion artifacts can degrade images. (3) Other artifacts and noise can reduce image quality. To tackle these problems, in this work, we propose a novel deep learning-based unsupervised method to estimate tag deformations on t-MRI images. The method has no annotation requirement during training, so with more training data are collected, our method can learn to predict more accurate cardiac deformation motion fields with minimal increased effort. 
In our method, we first track the  motion field in between two consecutive frames, using a bi-directional generative diffeomorphic registration network. 
Based on these initial motion field estimations, we then track the Lagrangian motion field between the reference frame and any other frame by a composition layer. 
The composition layer is differentiable, so it can update the learning parameters of the registration network with a global Lagrangian motion constraint, thus achieving a reasonable computation of motion fields. 

Our contributions could be summarized briefly as follows: 
(1) We propose a novel unsupervised method for t-MRI motion tracking, which can achieve a high accuracy of performance in a fast inference speed. 
(2) We propose a bi-directional diffeomorphic image registration network which could guarantee topology preservation and invertibility of the transformation, in which the likelihood of the warped image is modeled as a Boltzmann distribution, and a normalized cross correlation metric is incorporated in it, for its robust performance on image intensity time-variant registration problems.
(3) We propose a scheme to decompose the Lagrangian motion between the reference and any other frame into sums of consecutive frame motions and then improve the estimation of these motions by composing them back into the Lagrangian motion and posing a global motion constraint. 

\section{Background}
Regional myocardium motion quantification mainly focuses on the left ventricle (LV) myocardium (MYO) wall. It takes one t-MRI image sequence (usually a 2D video) as input and outputs a 2D motion field over time. The motion field is a 2D dense field depicting the non-rigid deformation of the LV MYO wall. The image sequence covers a full cardiac cycle. It starts from the end diastole (ED) phase, at which the ventricle begins to contract, then to the maximum contraction at end systole (ES) phase and back to relaxation to ED phase, as shown in Fig.~\ref{fig1}. Typically, we set a reference frame as the ED phase, and track the motion on any other later frame relative to the reference one. For t-MRI motion tracking, previous work was mainly based on phase, optical flow, and conventional non-rigid image registration.
\subsection{Phase-based Method}
Harmonic phase (HARP) based method is the most representative one for t-MRI image motion tracking~\cite{osman1999cardiac,osman2000imaging,liu2010shortest,liu2011incompressible,eldeeb2016accurate}. Periodic tags in the image domain correspond to spectral peaks in the Fourier domain of the image. Isolating the first harmonic peak region by a bandpass filter and performing an inverse Fourier transform of the selected region yields a complex harmonic image. The phase map of the complex image is the HARP image, which could be used for motion tracking since the harmonic phase of a material point is a time-invariant physics property, for simple translation. Thus, by tracking the harmonic phase vector of each pixel through time, one can track the position and, by extension, the displacement of each  pixel along time. 
However, due to cardiac motion, local variations of tag spacing and orientation at different frames may lead to erroneous phase estimation when using HARP, such as bifurcations in the reconstructed phase map, which also happens at boundaries and in large deformation regions of the myocardium~\cite{liu2010shortest}. Extending HARP, Gabor filters are used to refine phase map estimation by changing the filter parameters according to the local tag spacing and orientation, to automatically match different tag patterns in the image domain~\cite{chen2009automated, wang2008meshless, qian2011identifying}.

\subsection{Optical Flow Approach}
While HARP exploits specificity of quasiperiodic t-MRI, the optical flow (OF) based method is generic and can be applied to track objects in video sequences~\cite{horn1981determining, brox2010large,brox2004high,memin1998dense,wedel2009structure}. 
OF can estimate a dense motion field based on the basic assumption of image brightness constancy of local time-varying image regions with motion, at least for a very short time interval. The under-determined OF constraint equation is solved by variational principles in which some other regularization constraints are added in, including the image gradient, the phase or block matching. Although efforts have been made to seek more accurate regularization terms, OF approaches lack accuracy, especially for t-MRI motion tracking, due to the tag fading and large deformation problems~\cite{carranza2010motion, wang2019gradient}. More recently, convolutional neural networks (CNN) are trained to predict OF~\cite{dosovitskiy2015flownet, hur2019iterative, ilg2017flownet, lai2019bridging, liu2019selflow, ranjan2019competitive, meister2017unflow, sun2018pwc, yin2018geonet, wang2019unos, teed2020raft}. However, most of these works were supervised methods, with the need of a ground truth OF for training, which is nearly impossible to obtain for medical images.

\subsection{Image Registration-based Method}
Conventional non-rigid image registration methods have been used to estimate the deformation of the myocardium for a long time~\cite{shi2012comprehensive, rougon2005non, mcleod2011incompressible, chandrashekara2004analysis, morais2013cardiac, ledesma2008unsupervised}. Non-rigid registration schemes are formulated as an optimization procedure that maximizes a similarity criterion between the fixed image and the transformed moving image, to find the optimal transformation. Transformation models could be parametric models, including B-spline free-form deformation~\cite{shi2012comprehensive, morais2013cardiac, chandrashekara2004analysis}, and non-parametric models, including the variational method. Similarity criteria are generally chosen, such as mutual information and generalized information measures~\cite{rougon2005non}. All of these models are iteratively optimized, which is time consuming. 

Recently, deep learning-based methods have been applied to medical image registration and motion tracking. They are fast and have achieved at least comparable accuracy with conventional registration methods. Among those approaches, supervised methods~\cite{rohe2017svf} require ground truth deformation fields, which are usually synthetic. Registration accuracy thus will be limited by the quality of synthetic ground truth. Unsupervised methods~\cite{cao2017deformable, cao2018deformable, krebs2017robust, krebs2019learning, zhao2019recursive, de2017end, balakrishnan2018unsupervised, dalca2018unsupervised, niethammer2019metric, shen2019networks, shen2019region, mok2020fast} learn the deformation field by a loss function of the similarity between the fixed image and warped moving image. Unsupervised methods have been extended to cover deformable and diffeomorphic models. Deformable models~\cite{balakrishnan2018unsupervised, cao2017deformable, cao2018deformable} aim to learn the single directional deformation field from the fixed image to the moving image. Diffeomorphic models~\cite{dalca2018unsupervised, krebs2019learning,mok2020fast,shen2019region} learn the stationary velocity field (SVF) and integrate the SVF by a scaling and squaring layer, to get the diffeomorphic deformation field~\cite{dalca2018unsupervised}. A deformation field with diffeomorphism is differentiable and invertible, which ensures one-to-one mapping and preserves topology. Inspired by these works, we propose to use a bi-directional diffeomorphic registration network to track motions on t-MRI images.

\section{Method}
We propose an unsupervised learning method based on deep learning to track dense motion fields of objects that change over time. Although our method can be easily extended to other motion tracking tasks, without loss of generality, the design focus of the proposed method is t-MRI motion tracking.
\begin{figure}[t]
\begin{center}
\includegraphics[width=0.45\linewidth]{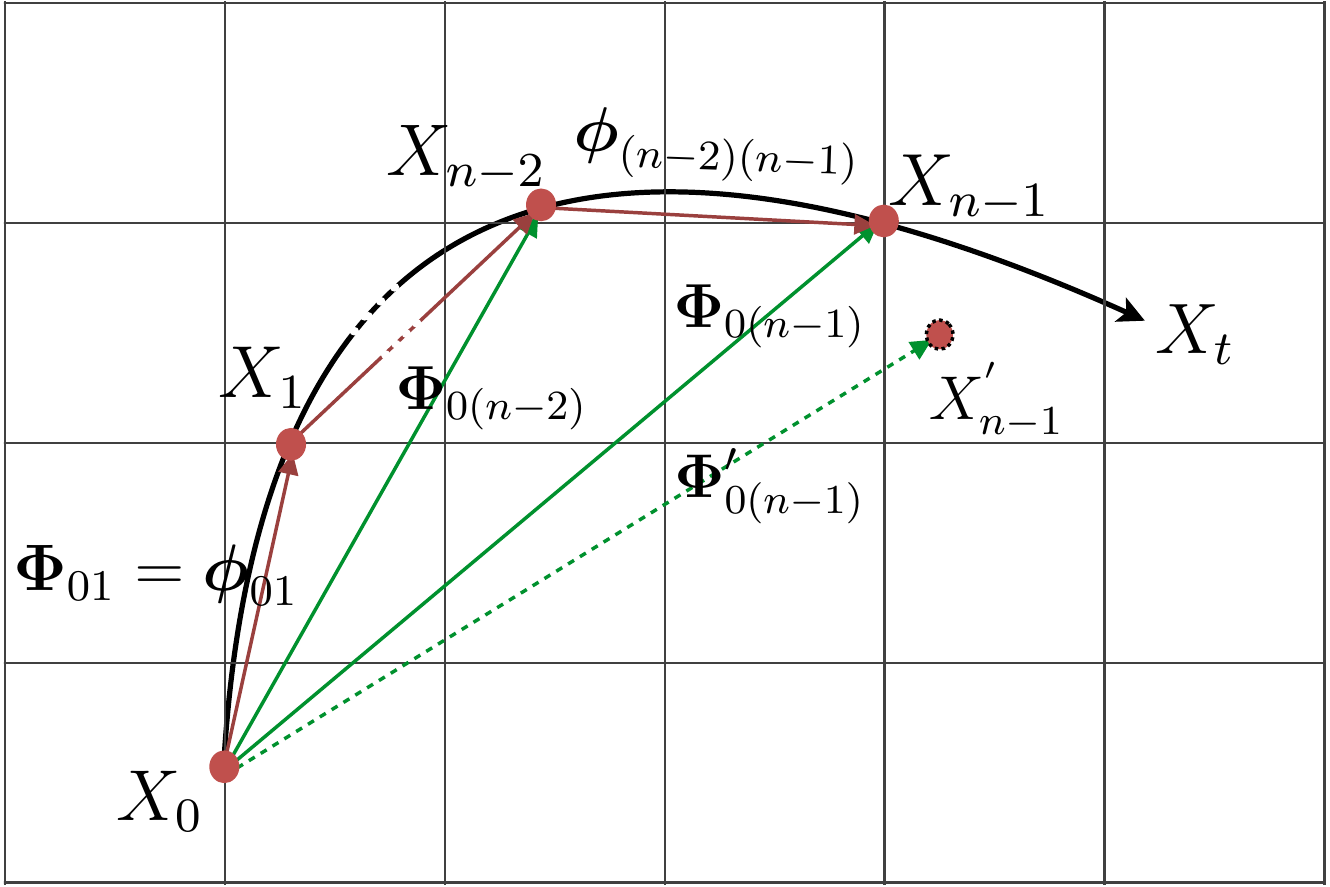}
\end{center}
   \caption{Interframe (INF) motion $\boldsymbol{\phi}$ and Lagrangian motion $\boldsymbol{\Phi}$.}
\label{fig2}
\end{figure}

\subsection{Motion Decomposition and Recomposition}
\label{sec31}
As shown in Fig.~\ref{fig2}, for a material point $m$ which moves from position $X_{0}$ at time $t_{0}$, we have its trajectory $X_{t}$. In a $N$ frames sequence, we only record the finite positions $X_{n} (n=0,1,...,N-1) $ of $m$. 
In a time interval $\Delta t=t_{n-1}-t_{n-2}$, the displacement can be shown pictorially as a vector $ \boldsymbol{\phi} _{(n-2)(n-1)}$, which in our work we call the interframe (INF) motion. 
A set of INF motions $ \left \{ \boldsymbol{\phi}_{t(t+1)} (t=0,1,...,n-2) \right \} $ will recompose the motion vector $ \boldsymbol{\Phi}_{0(n-1)}$, which we call the Lagrangian motion. 
While INF motion $\boldsymbol{\phi}_{t(t+1)}$ in between two consecutive frames is small if the time interval $\Delta t$ is small, net Lagrangian motion $\boldsymbol{\Phi}_{0(n-1)}$, however, could be very large in some frames of the sequence. 
For motion tracking, as we set the first frame as the reference frame, our task is to derive the Lagrangian motion $\boldsymbol{\Phi}_{0(n-1)}$ on any other later frame $t=n-1$. 
It is possible to directly track it based on the associated frame pairs, but for large motion, the tracking result $\boldsymbol{\Phi}'_{0(n-1)}$ could drift a lot. 
In a cardiac cycle, for a given frame $t=n-1$, since the amplitude $\parallel \boldsymbol{\phi}_{(n-2)(n-1)}\parallel $ $\leq$ $\parallel \boldsymbol{\Phi}_{0(n-1)}\parallel  $, decomposing $\boldsymbol{\Phi}_{0(n-1)}$ into $ \left \{ \boldsymbol{\phi}_{t(t+1)} (t=0,1,...,n-2) \right \} $, tracking $ \left \{ \boldsymbol{\phi}_{t(t+1)}\right \} $ at first, then composing them back to $\boldsymbol{\Phi}_{0(n-1)}$ will make sense. In this work, we follow this idea to obtain accurate motion tracking results on t-MRI images.

\begin{figure}[t]
\begin{center}
\includegraphics[width=0.8\linewidth]{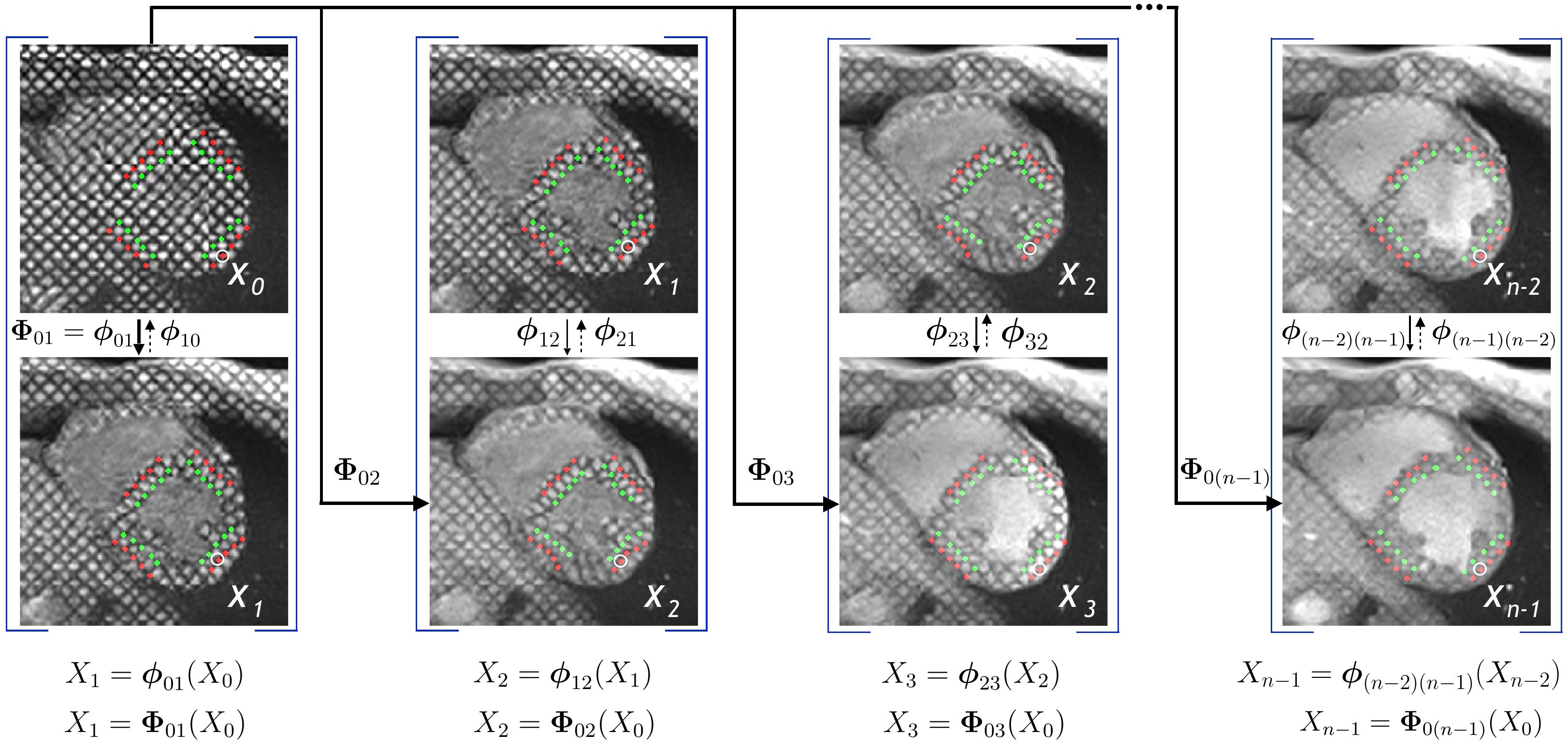}
\end{center}
   \caption{An overview of our scheme for regional myocardium motion tracking on t-MRI image sequences. $\boldsymbol{\phi} $: Interframe (INF) motion field between consecutive image pairs. $\boldsymbol{\Phi}$: Lagrangian motion field between the first frame and any other later frame.}
\label{fig3}
\end{figure}

\subsection{Motion Tracking on A Time Sequence}
\begin{figure}[t]
\begin{center}
\includegraphics[width=1.0\linewidth]{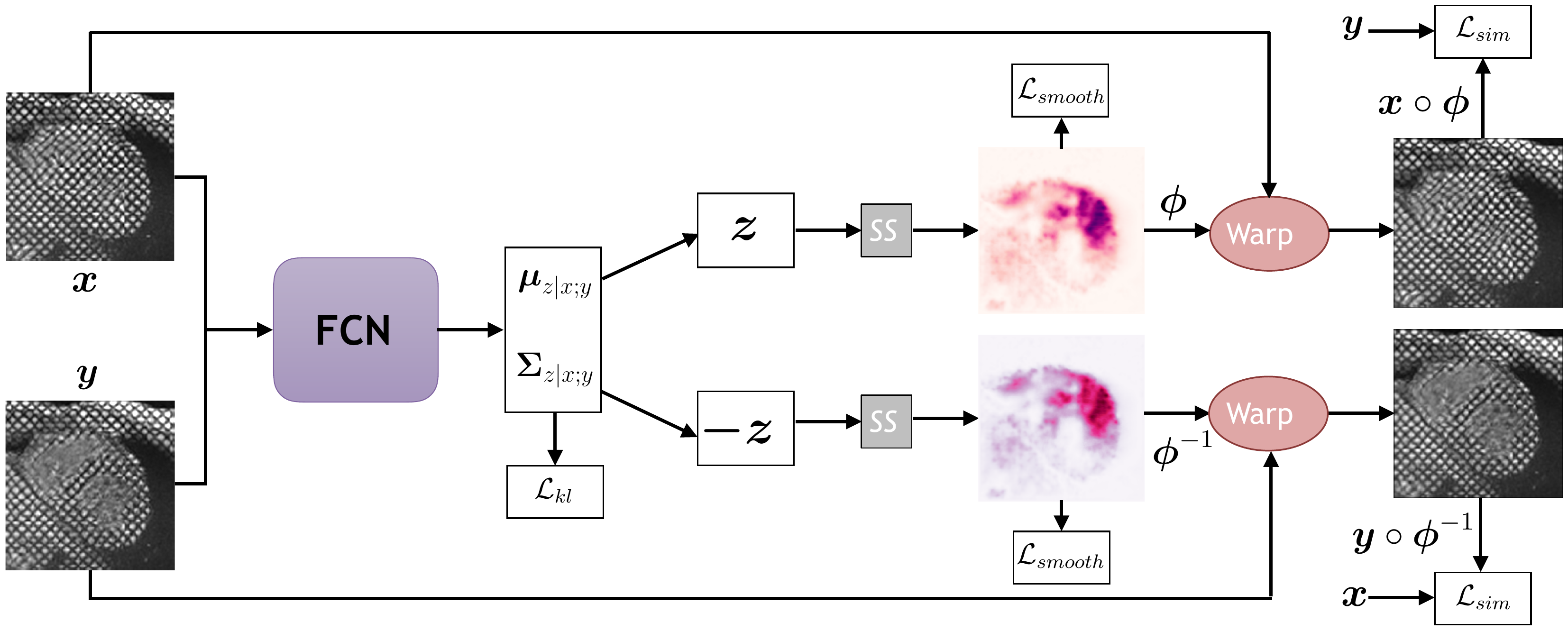}
\end{center}
   \caption{An overview of our proposed bi-directional forward-backward generative diffeomorphic registration network.}
\label{fig4}
\end{figure}

\begin{figure}[t]
\begin{center}
\includegraphics[width=0.7\linewidth]{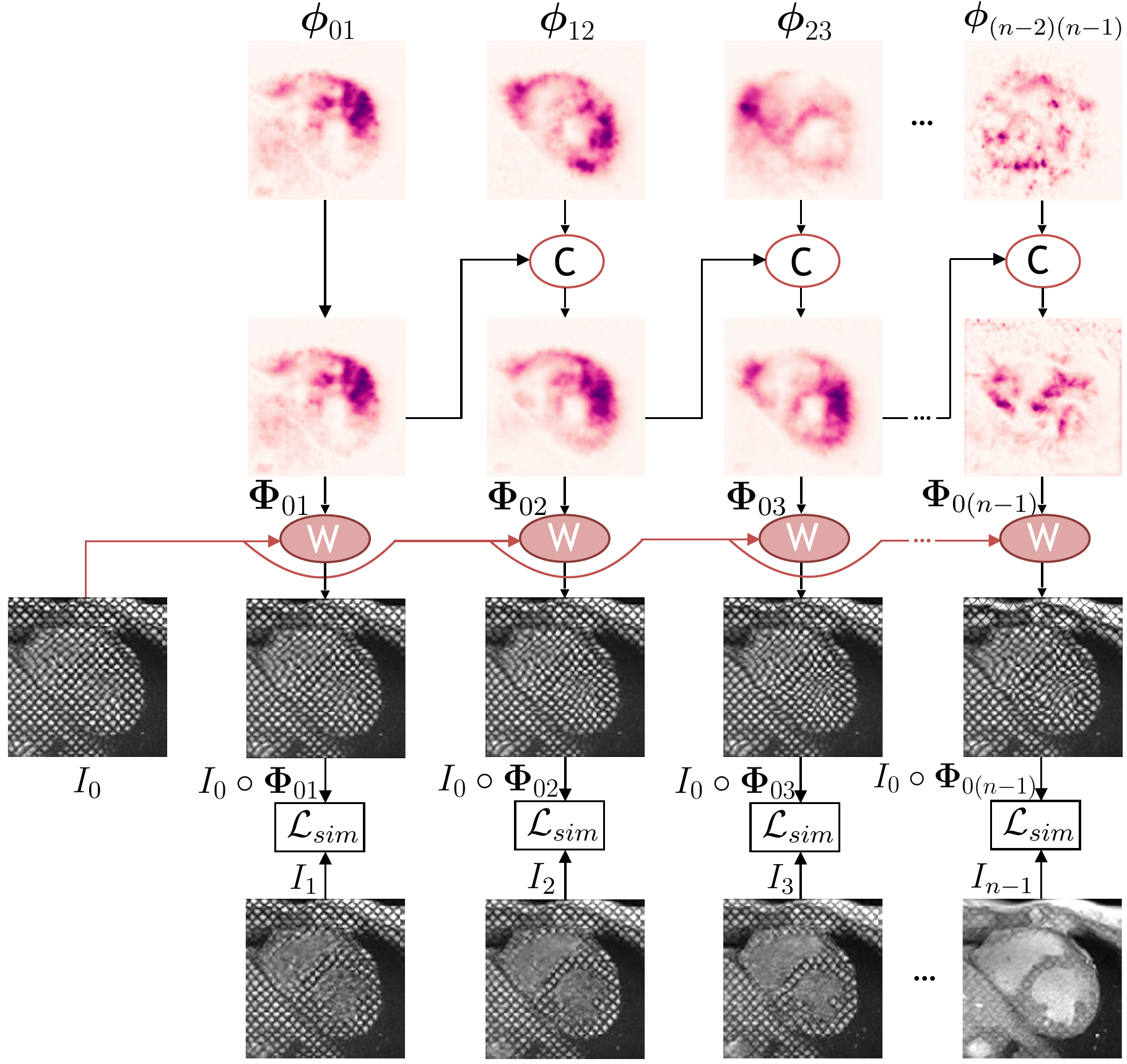}
\end{center}
   \caption{A composition layer $C$ that transforms INF motion field $\boldsymbol{\phi}$ to Lagrangian motion field $\boldsymbol{\Phi}$. ``W'' means ``warp''.} 
\label{fig5}
\end{figure}

\begin{figure}[t]
\begin{center}
\includegraphics[width=0.7\linewidth]{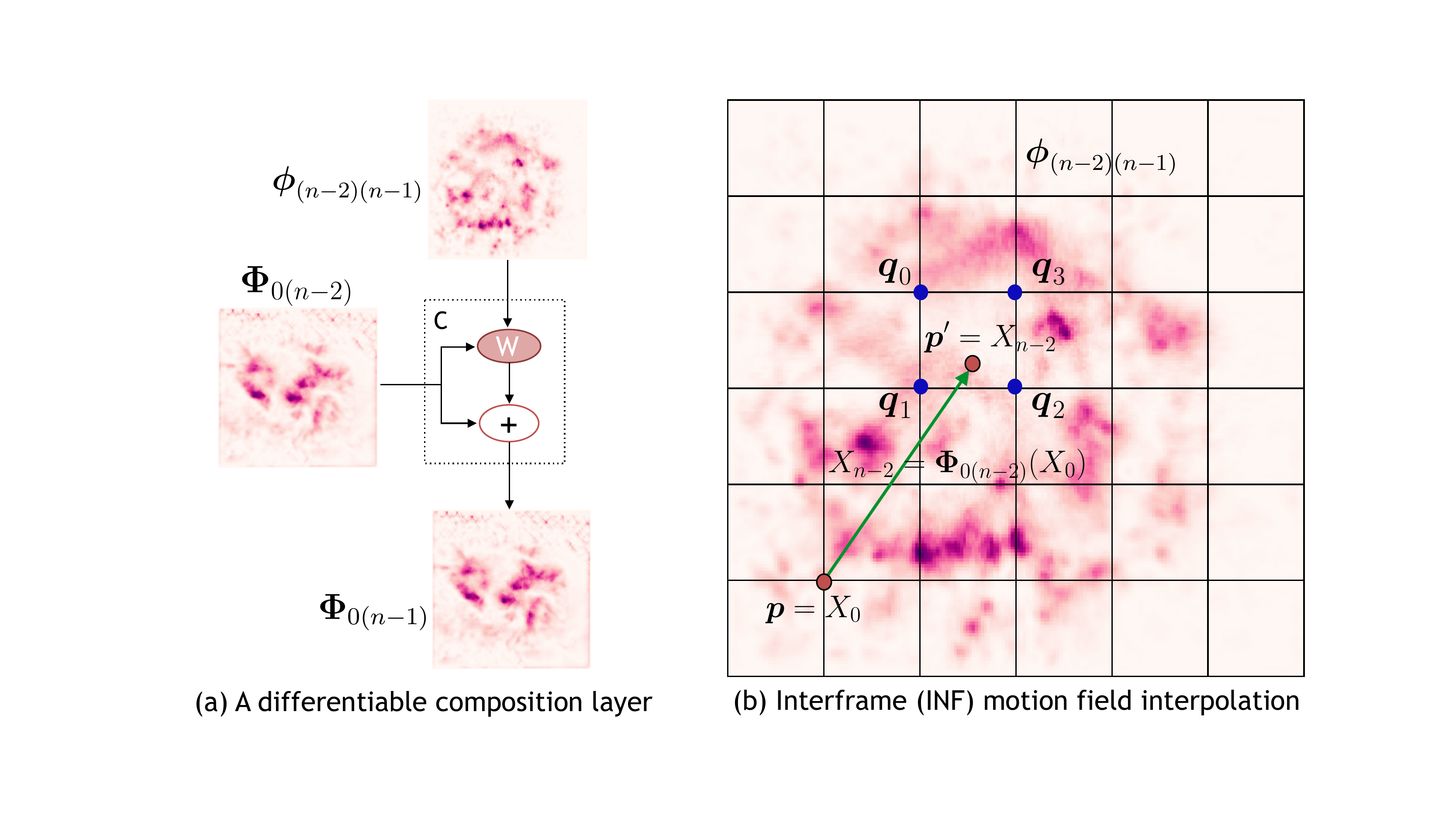}
\end{center}
   \caption{(a) The differentiable composition layer $C$. (b) INF motion field $\boldsymbol{\phi} $ interpolation at the new tracked position $\boldsymbol{p}'$.}
\label{fig6}
\end{figure}

Fig.~\ref{fig3} shows our scheme for myocardium motion tracking through time on a t-MRI image sequence. We first estimate the INF motion field $\boldsymbol{\phi}$ between two consecutive frames by a bi-directional diffeomorphic registration network, as shown in Fig.~\ref{fig4}. Once all the INF motion fields are obtained in the full time sequence, we compose them as the Lagrangian motion field $\boldsymbol{\Phi}$, which is shown in Fig.~\ref{fig5}. Motion tracking is achieved by predicting the position $X_{n-1}$ on an arbitrary frame moved from the position $X_{0}$ on the first frame with the estimated Lagrangian motion field: $X_{n-1}=\boldsymbol{\Phi} _{0(n-1)}(X_{0})$. In our method, motion composition is implemented by a differentiable composition layer $C$, as depicted in Fig.~\ref{fig6}. When training the registration network, such a differentiable layer can back-propagate the similarity loss between the warped reference image by Lagrangian motion field $\boldsymbol{\Phi}$ and any other later frame image as a global constraint and then update the parameters of the registration net, which in turn guarantees a reasonable INF motion field $\boldsymbol{\phi}$ estimation.

\subsection{Bi-Directional Forward-Backward Generative Diffeomorphic Registration Network}
As shown in Fig.~\ref{fig4}, we use a bi-directional forward-backward diffeomorphic registration network to estimate the INF motion field $\boldsymbol{\phi}$. Our network is modeled as a generative stochastic variational autoencoder (VAE)~\cite{kingma2013auto}. Let $\boldsymbol{x}$ and $\boldsymbol{y}$ be a 2D image pair, and let $\boldsymbol{z}$ be a latent variable that parameterizes the INF motion field $\boldsymbol{\phi}: \mathbb{R}^{2}\rightarrow \mathbb{R}^{2}$. Following the methodology of a VAE, we assume that the prior $p(\boldsymbol{z})$ is a multivariate Gaussian distribution with zero mean and covariance $\boldsymbol{\Sigma}_{z}$:
\begin{equation}
    p(\boldsymbol{z})\sim \mathcal N(\boldsymbol{z};\boldsymbol{0},\boldsymbol{\Sigma} _{z}) .
\end{equation}
The latent variable $\boldsymbol{z}$ could be applied to a wide range of representations for image registration. In our work, in order to obtain a diffeomorphism, we let $\boldsymbol{z}$ be a SVF which is generated as the path of diffeomorphic deformation field $\boldsymbol{\phi}^{(t)}$ parametrized by $t\in [0,1]$ as follows:
\begin{equation}
    \frac{d\boldsymbol{\phi}^{(t)}}{dt}=\boldsymbol{v}(\boldsymbol{\phi}^{(t)})=\boldsymbol{v}\circ \boldsymbol{\phi}^{(t)},
\end{equation}
where $\circ$ is a composition operator, $\boldsymbol{v}$ is the velocity field ($\boldsymbol{v} = \boldsymbol{z}$) and $\boldsymbol{\phi}^{(0)}=Id$ is an identity transformation. We follow ~\cite{arsigny2006log, ashburner2007fast, dalca2018unsupervised, mok2020fast} to integrate the SVF $\boldsymbol{v}$ over time $t=[0,1]$ by a scaling and squaring layer (SS) to obtain the final motion field $\boldsymbol{\phi}^{(1)}$ at time $t=1$. Specifically, starting from $ \boldsymbol{\phi}^{(1/2^{T})}=\boldsymbol{p}+\boldsymbol{v}(\boldsymbol{p})/2^{T}$ where $\boldsymbol{p}$ is a spatial location, by using the recurrence $ \boldsymbol{\phi}^{(1/2^{t})}=  \boldsymbol{\phi}^{(1/2^{t+1})}\circ \boldsymbol{\phi}^{(1/2^{t+1})}$ we can compute $ \boldsymbol{\phi}^{(1)}=  \boldsymbol{\phi}^{(1/2)}\circ \boldsymbol{\phi}^{(1/2)}$. In our experiments, $T=7$, which is chosen so that $\boldsymbol{v}(\boldsymbol{p})/2^{T}$ is small enough.
With the latent variable $\boldsymbol{z}$, we can compute the motion field $\boldsymbol{\phi}$ by the SS layer. We then use a spatial transform layer to warp image $\boldsymbol{x}$ by $\boldsymbol{\phi}$ and we obtain a noisy observation of the warped image, $\boldsymbol{x}\circ\boldsymbol{\phi}$, which could be a Gaussian distribution:
\begin{equation}
     p(\boldsymbol{y}|\boldsymbol{z};\boldsymbol{x})=\mathcal N(\boldsymbol{y};\boldsymbol{x}\circ\boldsymbol{\phi},\sigma^{2}\mathbb{I}),
\label{eq3}
\end{equation}
where $\boldsymbol{y}$ denotes the observation of warped image $\boldsymbol{x}$, $\sigma^{2}$ describes the variance of additive image noise. We call the process of warping image $\boldsymbol{x}$ towards $\boldsymbol{y}$ as the forward registration.

Our goal is to estimate the posterior probabilistic distribution $p(\boldsymbol{z}|\boldsymbol{y};\boldsymbol{x})$ for registration so that we obtain the most likely motion field $\boldsymbol{\phi}$ for a new image pair $(\boldsymbol{x},\boldsymbol{y})$ via maximum a posteriori estimation. However, directly computing this posterior is intractable. Alternatively, we can use a variational method, and introduce an approximate multivariate normal posterior probabilistic distribution $q_{\boldsymbol{\psi} }(\boldsymbol{z}|\boldsymbol{y};\boldsymbol{x})$ parametrized by a fully convolutional neural network (FCN) module $\boldsymbol{\psi}$ as:
\begin{equation}
    q_{\boldsymbol{\psi} }(\boldsymbol{z}|\boldsymbol{y};\boldsymbol{x})=\mathcal N(\boldsymbol{z};\boldsymbol{\mu}_{z|x,y},\boldsymbol{\Sigma }_{z|x,y}),
\end{equation}
where we let the FCN learn the mean $\boldsymbol{\mu}_{z|x,y}$ and diagonal covariance $\boldsymbol{\Sigma }_{z|x,y}$ of the posterior probabilistic distribution $q_{\boldsymbol{\psi} }(\boldsymbol{z}|\boldsymbol{y};\boldsymbol{x})$. When training the network, we implement a layer that samples a new latent variable $\boldsymbol{z}_{k}$ using the reparameterization trick: $\boldsymbol{z}_{k}=\boldsymbol{\mu}_{z|x,y}+\epsilon\boldsymbol{\Sigma }_{z|x,y}$, where $\epsilon \sim \mathcal N(\boldsymbol{0},\mathbb{I})$.

To learn parameters $\boldsymbol{\psi}$, we minimize the KL divergence between $q_{\boldsymbol{\psi} }(\boldsymbol{z}|\boldsymbol{y};\boldsymbol{x})$ and $p(\boldsymbol{z}|\boldsymbol{y};\boldsymbol{x})$, which leads to maximizing the evidence lower bound (ELBO)~\cite{kingma2013auto} of the log marginalized likelihood $log\ p(\boldsymbol{y}|\boldsymbol{x})$, as follows (detailed derivation in Supplementary Material):
\begin{equation}
\begin{aligned}
    &\underset{\boldsymbol{\psi}}{min}\, \, \mathcal{KL}[q_{\boldsymbol{\psi} }(\boldsymbol{z}|\boldsymbol{y};\boldsymbol{x})||p(\boldsymbol{z}|\boldsymbol{y};\boldsymbol{x})]\\ & 
    =\underset{\boldsymbol{\psi}}{min}\, \, \mathcal{KL}[q_{\boldsymbol{\psi} }(\boldsymbol{z}|\boldsymbol{y};\boldsymbol{x})||p(\boldsymbol{z})] -\mathbb{E}_{q}[log\, p(\boldsymbol{y}|\boldsymbol{z};\boldsymbol{x})] 
    \\ &+log\ p(\boldsymbol{y}|\boldsymbol{x}).
\end{aligned}
\label{eq5}
\end{equation}

In Eq.~(\ref{eq5}), the second term $-\mathbb{E}_{q}[log\, p(\boldsymbol{y}|\boldsymbol{z};\boldsymbol{x})]$ is called the reconstruction loss term in a VAE model. While we can model the distribution of $p(\boldsymbol{y}|\boldsymbol{z};\boldsymbol{x})$ as a Gaussian as in Eq.~(\ref{eq3}), which is equivalent to using a sum-of-squared difference (SSD) metric to measure the similarity between the warped image $\boldsymbol{x}$ and the observed $\boldsymbol{y}$,
in this work, we instead use a normalized local cross-correlation (NCC) metric, due to its robustness properties and superior results, especially for intensity time-variant image registration problems~\cite{avants2011reproducible, lorenzi2013lcc}. NCC of an image pair $I$ and $J$ is defined as:
\begin{equation}
\begin{aligned}
    &NCC(I, J)= \\&\sum_{\boldsymbol{p}\in \Omega }\frac{\left ( \sum_{\boldsymbol{p}_{i}\in\mathcal{W}}(I(\boldsymbol{p}_{i})-\bar{I}(\boldsymbol{p}))(J(\boldsymbol{p}_{i})-\bar{J}(\boldsymbol{p})) \right )^{2}}{\left (  \sum_{\boldsymbol{p}_{i}\in\mathcal{W}}(I(\boldsymbol{p}_{i})-\bar{I}(\boldsymbol{p}))^{2}\right )\left (\sum_{\boldsymbol{p}_{i}\in\mathcal{W}}(J(\boldsymbol{p}_{i})-\bar{J}(\boldsymbol{p}))^{2}\right )},
\end{aligned}
\end{equation}
where $\bar{I}(\boldsymbol{p})$ and $\bar{J}(\boldsymbol{p})$ are the local mean of $I$ and $J$ at position $\boldsymbol{p}$ respectively calculated in a $w^{2}$ window $\mathcal{W}$ centered at $\boldsymbol{p}$, $\Omega \subset \mathbb{R}^{2}$ is the 2D image spatial domain. In our experiments, we set $w=9$. A higher NCC indicates a better alignment, so the similarity loss between $I$ and $J$ could be: $\mathcal L_{sim}(I,J)=-NCC(I,J)$. Thus, we adopt the following Boltzmann distribution to model $p(\boldsymbol{y}|\boldsymbol{z};\boldsymbol{x})$ as:
\begin{equation}
    p(\boldsymbol{y}|\boldsymbol{z};\boldsymbol{x})\sim exp(-\gamma NCC(\boldsymbol{y}, \boldsymbol{x}\circ\boldsymbol{\phi})),
\end{equation}
where $\gamma$ is a negative scalar hyperparameter. Finally, we formulate the loss function as:
\begin{equation}
\begin{aligned}
    \mathcal L_{kl}&=\mathcal{KL}[q_{\boldsymbol{\psi} }(\boldsymbol{z}|\boldsymbol{y};\boldsymbol{x})||p(\boldsymbol{z})]-\mathbb{E}_{q}[log\, p(\boldsymbol{y}|\boldsymbol{z};\boldsymbol{x})]+const
    \\&=\frac{1}{2}\left [ tr(\lambda D\boldsymbol{\Sigma }_{z|x,y}-log\boldsymbol{\Sigma }_{z|x,y}) +\boldsymbol{\mu}_{z|x,y}^{T}\boldsymbol{\Lambda }_{z}\boldsymbol{\mu}_{z|x,y}\right ]
    \\&+\frac{\gamma}{K} \underset{k}{\sum }NCC(\boldsymbol{y},\boldsymbol{x}\circ\boldsymbol{\phi}_{k})+const,
\end{aligned}
\label{eq9}
\end{equation}
where $\boldsymbol{D}$ is the graph degree matrix defined on the 2D image pixel grid and $K$ is the number of samples used to approximate the expectation, with $K=1$ in our experiments. We let $\boldsymbol{L}=\boldsymbol{D}-\boldsymbol{A}$ be the Laplacian of a neighborhood graph defined on the pixel grid, where $\boldsymbol{A}$ is a pixel neighborhood adjacency matrix. To encourage the spatial smoothness of SVF $\boldsymbol{z}$, we set $\boldsymbol{\Lambda }_{z}=\boldsymbol{\Sigma }_{z}^{-1}= \lambda\boldsymbol{L}$~\cite{dalca2018unsupervised}, where $\lambda$ is a parameter controlling the scale of the SVF $\boldsymbol{z}$.

With the SVF representation, we can also compute an inverse motion field $\boldsymbol{\phi}^{-1}$ by inputting $-\boldsymbol{z}$ into the SS layer: $\boldsymbol{\phi}^{-1}=SS(-\boldsymbol{z})$. Thus we can warp image $\boldsymbol{y}$ towards image $\boldsymbol{x}$ (the backward registration) and get the observation distribution of warped image $\boldsymbol{y}$: $p(\boldsymbol{x}|\boldsymbol{z};\boldsymbol{y})$. We minimize the KL divergence between $q_{\boldsymbol{\psi} }(\boldsymbol{z}|\boldsymbol{x};\boldsymbol{y})$ and $p(\boldsymbol{z}|\boldsymbol{x};\boldsymbol{y})$ which leads to maximizing the ELBO of the log marginalized likelihood $log\ p(\boldsymbol{x}|\boldsymbol{y})$ (see supplementary material for detailed derivation).
In this way, we can add the backward KL loss term into the forward KL loss term and get:
\begin{equation}
\begin{aligned}
    &\mathcal L_{kl}(\boldsymbol{x},\boldsymbol{y})=
    \\&\mathcal{KL}[q_{\boldsymbol{\psi} }(\boldsymbol{z}|\boldsymbol{y};\boldsymbol{x})||p(\boldsymbol{z}|\boldsymbol{y};\boldsymbol{x})]+\mathcal{KL}[q_{\boldsymbol{\psi} }(\boldsymbol{z}|\boldsymbol{x};\boldsymbol{y})||p(\boldsymbol{z}|\boldsymbol{x};\boldsymbol{y})]
    \\&=\mathcal{KL}[q_{\boldsymbol{\psi} }(\boldsymbol{z}|\boldsymbol{y};\boldsymbol{x})||p(\boldsymbol{z})]-\mathbb{E}_{q}[log\, p(\boldsymbol{y}|\boldsymbol{z};\boldsymbol{x})]+
    \\&\mathcal{KL}[q_{\boldsymbol{\psi} }(\boldsymbol{z}|\boldsymbol{x};\boldsymbol{y})||p(\boldsymbol{z})]-\mathbb{E}_{q}[log\, p(\boldsymbol{x}|\boldsymbol{z};\boldsymbol{y})]+const
    \\&=tr(\lambda D\boldsymbol{\Sigma }_{z|x,y}-log\boldsymbol{\Sigma }_{z|x,y}) +\boldsymbol{\mu}_{z|x,y}^{T}\boldsymbol{\Lambda }_{z}\boldsymbol{\mu}_{z|x,y}+
    \\&\frac{\gamma}{K}\sum_{k}(NCC(\boldsymbol{y},\boldsymbol{x}\circ\boldsymbol{\phi}_{k})+NCC(\boldsymbol{x},\boldsymbol{y}\circ\boldsymbol{\phi}_{k}^{-1}))+const.
\end{aligned}
\end{equation}
The second term spatially smooths the mean $\boldsymbol{\mu_{z|x,y}}$, as we can expand it as $\boldsymbol{\mu}_{z|x,y}^{T}\boldsymbol{\Lambda }_{z}\boldsymbol{\mu}_{z|x,y}=\frac{\lambda}{2}\sum \sum_{j\in N(i)}(\boldsymbol{\mu}[i]-\boldsymbol{\mu}[j])^{2}$, where $N(i)$ are the neighbors of pixel $i$. While this is an implicit smoothness of the motion field, we also enforce the explicit smoothness of the motion field $\boldsymbol{\phi}$ by penalizing its gradients: $\mathcal L_{smooth}(\boldsymbol{\phi})=\left \| \bigtriangledown \boldsymbol{\phi} \right \|_{2}^{2}$.

Such a bi-directional registration architecture not only enforces the invertibility of the estimated motion field but also provides a path for the inverse consistency of the predicted motion field. Since the tags fade in later frames in a cardiac cycle and there exists a through-plane motion problem, we need this forward-backward constraint to obtain a more reasonable motion tracking result.

\subsection{Global Lagrangian Motion Constraints}
\label{sec34}
After we get all the INF motion fields in a t-MRI image sequence, we design a differentiable composition layer $C$ to recompose them as the Lagrangian motion field $\boldsymbol{\Phi}$, as shown in Fig.~\ref{fig5}. From Fig.~\ref{fig2} we can get, $\boldsymbol{\Phi}_{01}=\boldsymbol{\phi}_{01}$, $\boldsymbol{\Phi}_{0(n-1)}=\boldsymbol{\Phi}_{0(n-2)}+\boldsymbol{\phi}_{(n-2)(n-1)} (n>2)$. However, as Fig.~\ref{fig6} (b) shows, the new position $\boldsymbol{p}'=X_{n-2}=\boldsymbol{\Phi}_{0(n-2)}(X_{0})$ could be a sub-pixel location, and because INF motion field values are only defined at integer locations, we linearly interpolate the values between the four neighboring pixels:
\begin{equation}
\begin{aligned}
    & \boldsymbol{\phi}_{(n-2)(n-1)}\circ\boldsymbol{\Phi}_{0(n-2)}(X_{0}) \\&=\sum_{\boldsymbol{q}\in N(\boldsymbol{p}')}\boldsymbol{\phi}_{(n-2)(n-1)}[\boldsymbol{q}]\prod_{d\in\left \{ x,y \right \}}(1-\left | \boldsymbol{p}'_{d}-\boldsymbol{q}_{d} \right |),
\end{aligned}
\label{eq12}
\end{equation}
where $N(\boldsymbol{p}')$ are the pixel neighbors of $\boldsymbol{p}'$, and $d$ iterates over dimensions of the motion field spatial domain. Note here we use $\boldsymbol{\phi}[\cdot ]$ to denote the values of $\boldsymbol{\phi}$ at location $[\cdot]$ to differentiate it from $\boldsymbol{\phi}(\cdot)$, which means a mapping that moves one location $X_{n-2}$ to another $X_{n-1}$; the same is used with $\boldsymbol{\Phi}[\cdot]$ in the following. In this formulation, we use a spatial transform layer to implement the INF motion field interpolation. Then we add the interpolated $\boldsymbol{\phi}_{(n-2)(n-1)}$ to the $\boldsymbol{\Phi}_{0(n-2)}$ and get the $\boldsymbol{\Phi}_{0(n-1)} (n>2)$, as shown in Fig.~\ref{fig6} (a) (see details of computing $\boldsymbol{\Phi}$ from $\boldsymbol{\phi}$ in Algorithm~\ref{alg1} in Supplementary Material).

With the Lagrangian motion field $\boldsymbol{\Phi}_{0(n-1)}$, we can warp the reference frame image $I_{0}$ to any other frame at $t=n-1$: $I_{0}\circ\boldsymbol{\Phi}_{0(n-1)}$. By measuring the NCC similarity between $I_{n-1}$ and $I_{0}\circ\boldsymbol{\Phi}_{0(n-1)}$, we form a global Lagrangian motion consistency constraint:
\begin{equation}
    \mathcal L_{g}=-\sum_{n=2}^{N}NCC(I_{n-1},I_{0}\circ\boldsymbol{\Phi}_{0(n-1)}),
\end{equation}
where $N$ is the total frame number of a t-MRI image sequence. This global constraint is necessary to guarantee that the estimated INF motion field $\boldsymbol{\phi}$ is reasonable to satisfy a global Lagrangian motion field. Since the INF motion estimation could be erroneous, especially for large motion in between two consecutive frames, the global constraint can correct the local estimation within a much broader horizon by utilizing temporal information. Further, we also enforce the explicit smoothness of the Lagrangian motion field $\boldsymbol{\Phi}$ by penalizing its gradients: $\mathcal L_{smooth}(\boldsymbol{\Phi})=\left \| \bigtriangledown \boldsymbol{\Phi} \right \|_{2}^{2}$.

To sum up, the complete loss function of our model is the weighted sum of $\mathcal L_{kl}$, $\mathcal L_{smooth}$ and $\mathcal L_{g}$:
\begin{equation}
\begin{aligned}
    &\mathcal L=\sum_{n=0}^{N-2}[\mathcal L_{kl}(I_{n},I_{n+1})+\alpha_{1}( \mathcal L_{smooth}(\boldsymbol{\phi}_{n(n+1)})+
    \\&\mathcal L_{smooth}(\boldsymbol{\phi}_{(n+1)n}))
    +\alpha_{2} \mathcal L_{smooth}(\boldsymbol{\Phi}_{0(n+1)})]+\beta \mathcal L_{g},
\end{aligned}
\end{equation}
where $\alpha_{1}$, $\alpha_{2}$ and $\beta$ are the weights to balance the contribution of each loss term.

\section{Experiments}
\subsection{Dataset and Pre-Processing}
To evaluate our method, we used a clinical t-MRI dataset which consists of 23 subjects' whole heart scans. Each scan set covers the 2-, 3-, 4-chamber and short-axis (SAX) views. For the SAX views, it includes several slices starting from the base to the apex of the heart ventricle; each set has approximately 10 2D slices, each of which covers a full cardiac cycle forming a 2D sequence. In total, there are 230 2D sequences in our dataset. For each sequence, the frame numbers vary from $16\sim25$. We first extracted the region of interest (ROI) from the images to cover the heart, then resampled them to the same in-plane spatial size $192\times192$. Each sequence was used as input to the model to track the cyclic cardiac motion. For the temporal dimension, if the frames are less than 25, we copy the last frame to fill the gap. So each input data is a 2D sequence consists of $25$ frames whose spatial resolution is $192\times192$. We randomly split the dataset into $140$, $30$ and $60$ sequences as the train, validation and test sets, respectively (Each set comes from different subjects). For each 2D image, we normalized the image values by first dividing them with the 2 times of median intensity value of the image and then truncating the values to be $[0, 1]$. We also did $40$ times data augmentation with random rotation, translation, scaling and Gaussian noise addition. 

\subsection{Evaluation Metrics}
Two clinical experts annotated $8\sim32$ landmarks on the LV MYO wall for each testing sequence, for example, as shown in Fig.~\ref{fig7} by the red dots; they double checked all the annotations carefully. During evaluation, we input the landmarks on the first frame and predicted their locations on the later frames by the Lagrangian motion field $\boldsymbol{\Phi}$. Following the metric used in~\cite{chandrashekara2004analysis}, we used the root mean squared (RMS) error of distance between the centers of predicted landmark ${X}'$ and ground truth landmark $X$ to assess motion tracking accuracy. In addition, we evaluated the diffeomorphic property of the predicted INF motion field $\boldsymbol{\phi}$, using the Jacobian determinant $det(J_{\boldsymbol{\phi}}(\boldsymbol{p}))$  (detailed definitions of the two metrics in Supplementary Material).

\subsection{Baseline Methods}
We compared our proposed method with two conventional t-MRI motion tracking methods. The first one is HARP~\cite{osman1999cardiac}. We reimplemented it in MATLAB (R2019a). Another one is the variational OF method\footnote{Code is online \url{http://www.iv.optica.csic.es/page49/page54/page54.html}}~\cite{carranza2010motion}, which uses a total variation (TV) regularization term. 
We also compared our method with the unsupervised deep learning-based medical image registration methods VM~\cite{balakrishnan2018unsupervised} and VM-DIF~\cite{dalca2018unsupervised}, which are recent cutting-edge unsupervised image registration approaches. VM uses SSD (MSE) or NCC loss for training, while VM-DIF uses SSD loss. We used their official implementation code online\footnote{\url{https://github.com/voxelmorph/voxelmorph}}, and trained VM and VM-DIF from scratch by following the optimal hyper-parameters suggested by the authors.

\subsection{Implementation Details}
We implemented our method with Pytorch. For the FCN, the architecture is the same as in~\cite{dalca2018unsupervised}. We used the Adam optimizer with a learning rate of $5e^{-4}$ to train our model. For the hyper-parameters, we set $\alpha_{1}=5$, $\alpha_{2}=1$, $\beta=0.5$, $\gamma=-0.5$, $\lambda=10$, via grid search. All models were trained on an NVIDIA Quadro RTX 8000 GPU. The models with the lowest loss on the validation set were selected for evaluation. 

\begin{table}[t]
\begin{center}
\resizebox{1.0\columnwidth}{!}{
\begin{tabular}{l|c|c|c}
\hline
Method & RMS ($mm$) $\downarrow$& $det(J_{\boldsymbol{\phi}}) \leqslant  0$ $(\#)$ $\downarrow$& Time ($s$) $\downarrow$\\
\hline\hline
HARP &$3.814\pm1.098$  &$5950.4\pm1709.4$ &$124.3446\pm21.0055$\\
OF-TV &$2.529\pm0.726$ &$80.3\pm71.1$ &$36.6764\pm10.6163$\\
VM (SSD) &$3.799\pm1.031$ &$622.2\pm390.7$ &$\mathbf{0.0161}\pm0.0355$\\
VM (NCC) &$2.856\pm1.185$ &$11.4\pm12.3$ &$0.0162\pm0.0351$\\
VM-DIF &$3.235\pm1.144$ &$1.7\pm1.6$ &$0.0202\pm\mathbf{0.0332}$\\
Ours &$\mathbf{1.628\pm0.587}$ &$\mathbf{0.0\pm0.0}$ &$0.0202\pm0.0339$\\
\hline
\end{tabular}}
\end{center}
\caption{Average RMS error, number of pixels with non-positive Jacobian determinant and running time.}
\label{table1}
\end{table}

\subsection{Results}
\subsubsection{Motion Tracking Performance}
In Table~\ref{table1}, we show the average RMS error and the number of pixels with non-positive Jacobian determinant for baseline motion tracking methods and ours. We also show an example in Fig.~\ref{fig7} (full sequence results in Supplementary Material). Mean and standard deviation of the RMS errors across a cardiac cycle are shown in Fig.~\ref{fig8}.
For HARP, which is based on phase estimation, there could be missing landmark tracking results on the septal wall, due to unrealistic phase estimations, as indicated by the arrows in Fig.~\ref{fig7}. In addition, depending on the accuracy of the phase estimation, the tracked landmarks could drift far away although the points of each landmark should be spatially close. 
OF-TV performs better than HARP, but it suffers from tag fading and large motion problems. The tracking results drifted a lot in the later frames. As shown in Fig.~\ref{fig8}, the RMS error for OF-TV increased with the cardiac cycle phase.
VM (NCC) is better than VM (SSD), because of the robustness of NCC loss for intensity time-variant image registration. While VM-DIF uses the SSD loss, it is better than VM (SSD) because of the diffeomorphic motion field that VM-DIF aims to learn. However, VM-DIF is worse than VM (NCC), indicating that NCC loss is more suitable for intensity time-variant image registration problems than SSD loss.  
VM and VM-DIF are worse than OF-TV, which suggests that we cannot apply the cutting-edge unsupervised registration methods to the t-MRI motion tracking problem without any adaptation.
Our method obtains the best performance since it utilizes the NCC loss, bi-directional and global Lagrangian constraints, as well as the diffeomorphic nature of the learned motion field. The diffeomorphic attribute is also reflected by the Jacobian determinant. Our method maintains the number of pixels with non-positive Jacobian determinant as zero, which indicates the learned motion field is smooth, topology preserving and ensures one-to-one mapping. 

\begin{figure}[t]
\begin{center}
\includegraphics[width=0.85\linewidth]{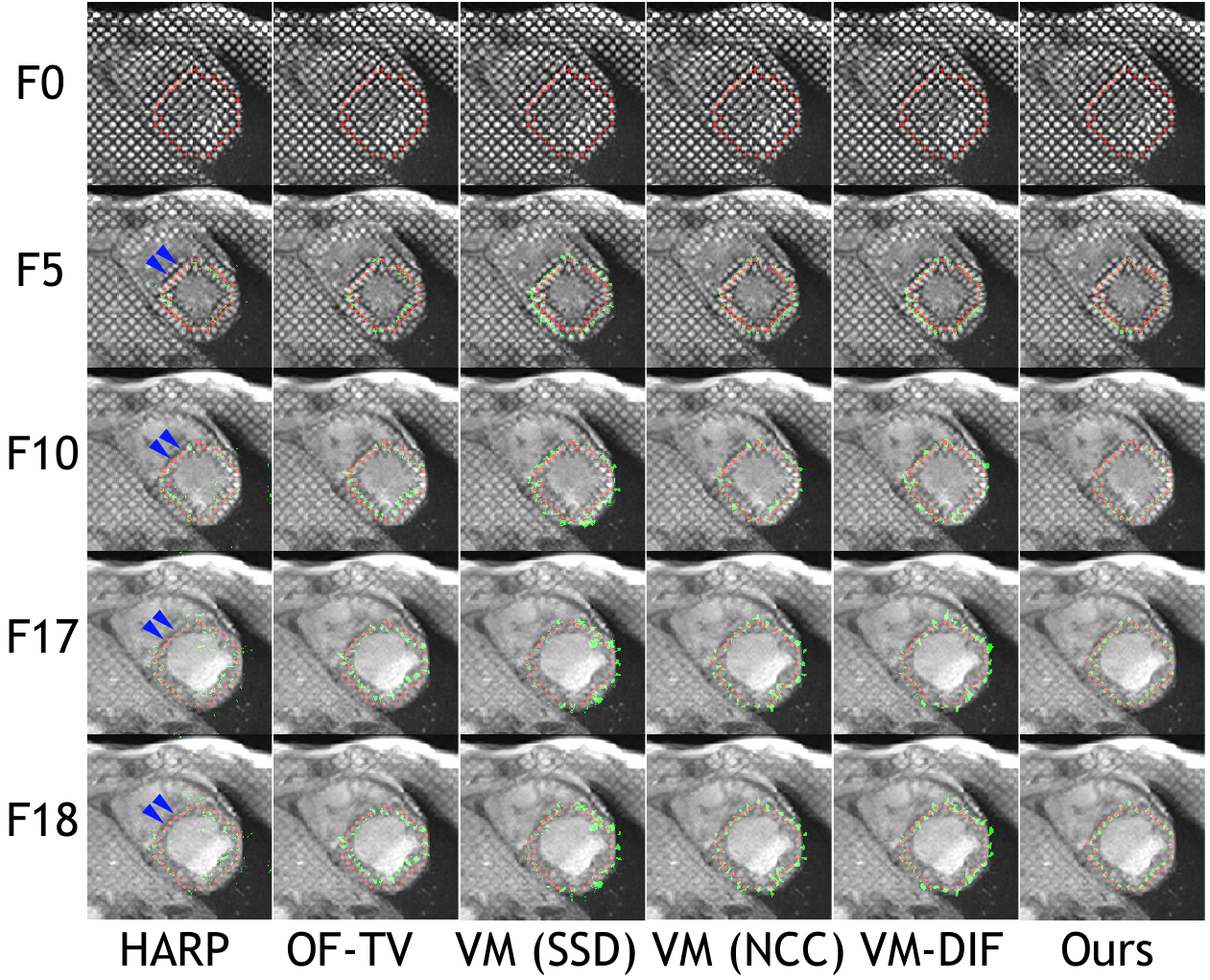}
\end{center}
   \caption{Motion tracking results on a t-MRI image sequence of 19 frames (best viewed zoomed in). Red is ground truth, green is prediction. ``F'' means ``frame''.}
\label{fig7}
\end{figure}

\begin{figure}[t]
\begin{center}
\includegraphics[width=0.5\linewidth]{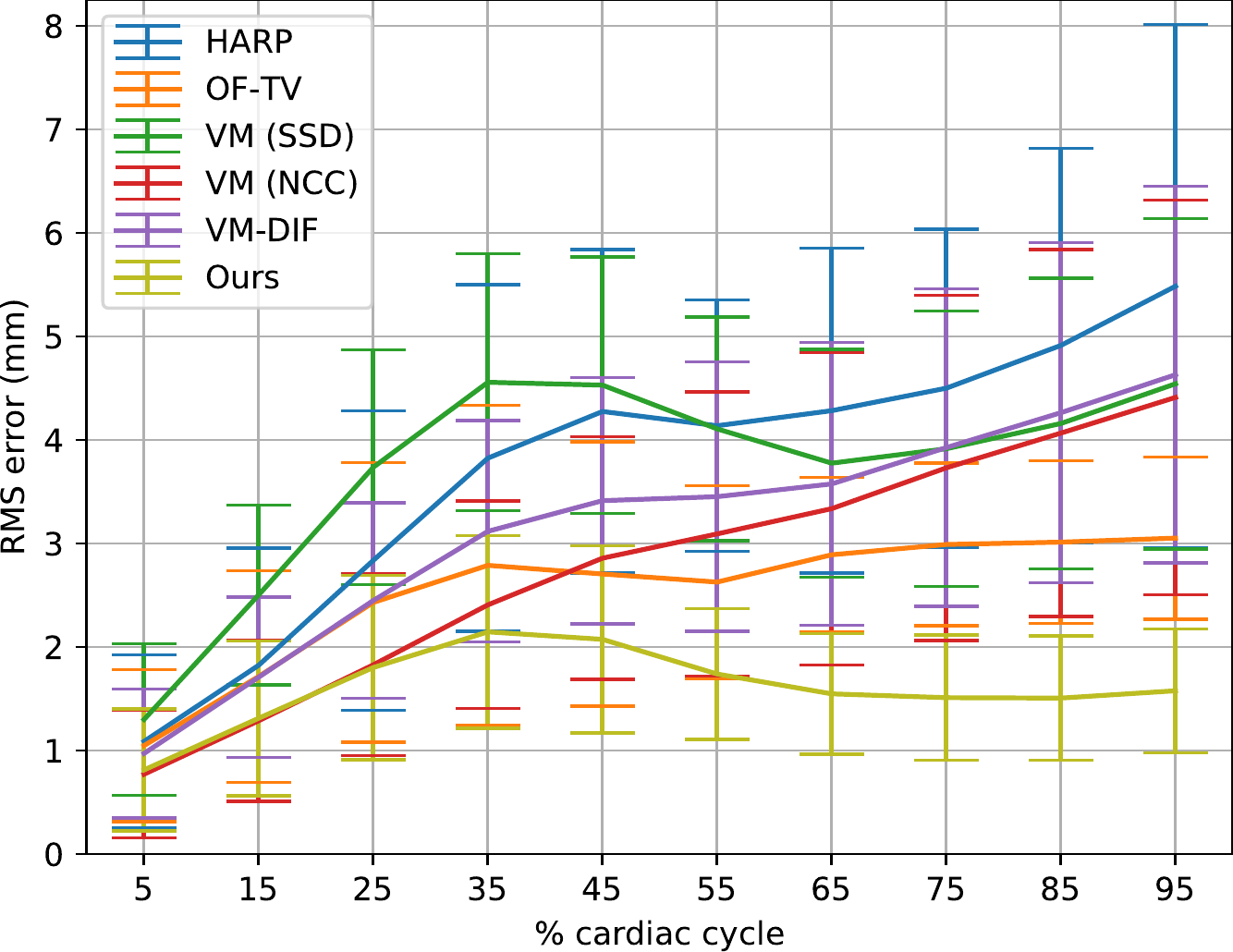}
\end{center}
   \caption{Mean and standard deviation of the RMS errors across a cardiac cycle for baseline methods and ours.}
\label{fig8}
\end{figure}

\subsubsection{Ablation Study and Results}
To compare the efficiency of tracking Lagrangian motion and INF motion, we designed two kinds of restricted models. One is to do registration between the reference and any other later frame, the other is registration between consecutive frames: A1 (forward Lagrangian tracking) and A2 (forward INF tracking).
To explore the effect of bi-directional regularization, we studied the forward-backward model: A3 (A2 + backward INF tracking). 
We then studied the effect of explicit smoothness over the INF motion field: A4 (A3 + INF motion field $\boldsymbol{\phi}$ smooth). 
To validate our proposed global Lagrangian motion constraint, we studied models with every four frames and with full sequence global constraint: A5 (A4 + every 4 frames Lagrangian constraint) and A6 (A4 + full sequence Lagrangian constraint).
We also studied the effect of explicit smoothness over the Lagrangian motion field: Ours (A6 + Lagrangian motion field $\boldsymbol{\Phi}$ smooth).

In Table~\ref{table2}, we show the average RMS error and number of pixels with non-positive Jacobian determinant. We also show an example in Fig.~\ref{fig9} (full sequence results in Supplementary Material). The mean and standard devation of RMS errors for each model across a cardiac cycle is shown in Fig.~\ref{fig10}. 
As we previously analyzed in Section~\ref{sec31}, directly tracking Lagrangian motion will deduce a drifted result for large motion frames, as shown in frame $5\sim11$ for A1 in Fig.~\ref{fig9}. 
Although forward-only INF motion tracking (A2) performs worse than A1 on average, mainly due to tag fading  on later frames, bi-directional INF motion tracking (A3) is better than both A1 and A2. From Fig.~\ref{fig10}, A3 mainly improves the performance of INF motion tracking estimation on later frames with the help of inverse consistency of the backward constraint. 
The explicit INF and Lagrangian motion field smoothness regularization (A4 and ours) helps to smooth the learned motion field for later frames with the prior that spatially neighboring pixels should move smoothly together. However, the smoothness constraints make it worse for the earlier (systolic) frames, which warrants a further study of a time-variant motion field smoothness constraint in the future.
Our proposed global Lagrangian motion constraint greatly improved the estimation of the large INF motion  (A6 and ours). As shown in Fig.~\ref{fig9}, beginning with frame $9$, the heart gets into the rapid early filling phase. INF motion in between frame $9$ and $10$ is so large that, without a global motion constraint (A3 and A4), the tracking results would drift a lot on the lateral wall as indicated by arrows. What's worse, such a drift error will accumulate over the following frames, which results in erroneous motion estimation on a series of frames. The proposed global constraint, however, could correct such an unreasonable INF motion estimation and a full sequence global constraint (A6) achieves better results than the segmented every $4$ frames constraint (A5).   
All models have no non-positive Jacobian determinants, suggesting that the learned motion fields guarantee one-to-one mapping.

\begin{table}
\begin{center}
\resizebox{0.65\columnwidth}{!}{
\begin{tabular}{l|c|c}
\hline
Model & RMS ($mm$) $\downarrow$ &  $det(J_{\boldsymbol{\phi}}) \leqslant  0$ $(\#)$ $\downarrow$ \\
\hline\hline
A1  &$2.958\pm0.695$  &$0.0\pm0.0$ \\
A2  &$2.977\pm1.217$  &$0.0\pm0.0$ \\
A3  &$1.644\pm0.611$  &$0.0\pm0.0$\\
A4  &$1.654\pm\mathbf{0.586}$ &$0.0\pm0.0$ \\
A5  &$1.704\pm0.677$  &$0.0\pm0.0$ \\
A6  &$1.641\pm0.637$  &$0.0\pm0.0$\\
Ours  &$\mathbf{1.628}\pm0.587$ &$\mathbf{0.0\pm0.0}$ \\
\hline
\end{tabular}}
\end{center}
\caption{Ablation study results.}
\label{table2}
\end{table}

\begin{figure}[t]
\begin{center}
\includegraphics[width=0.85\linewidth]{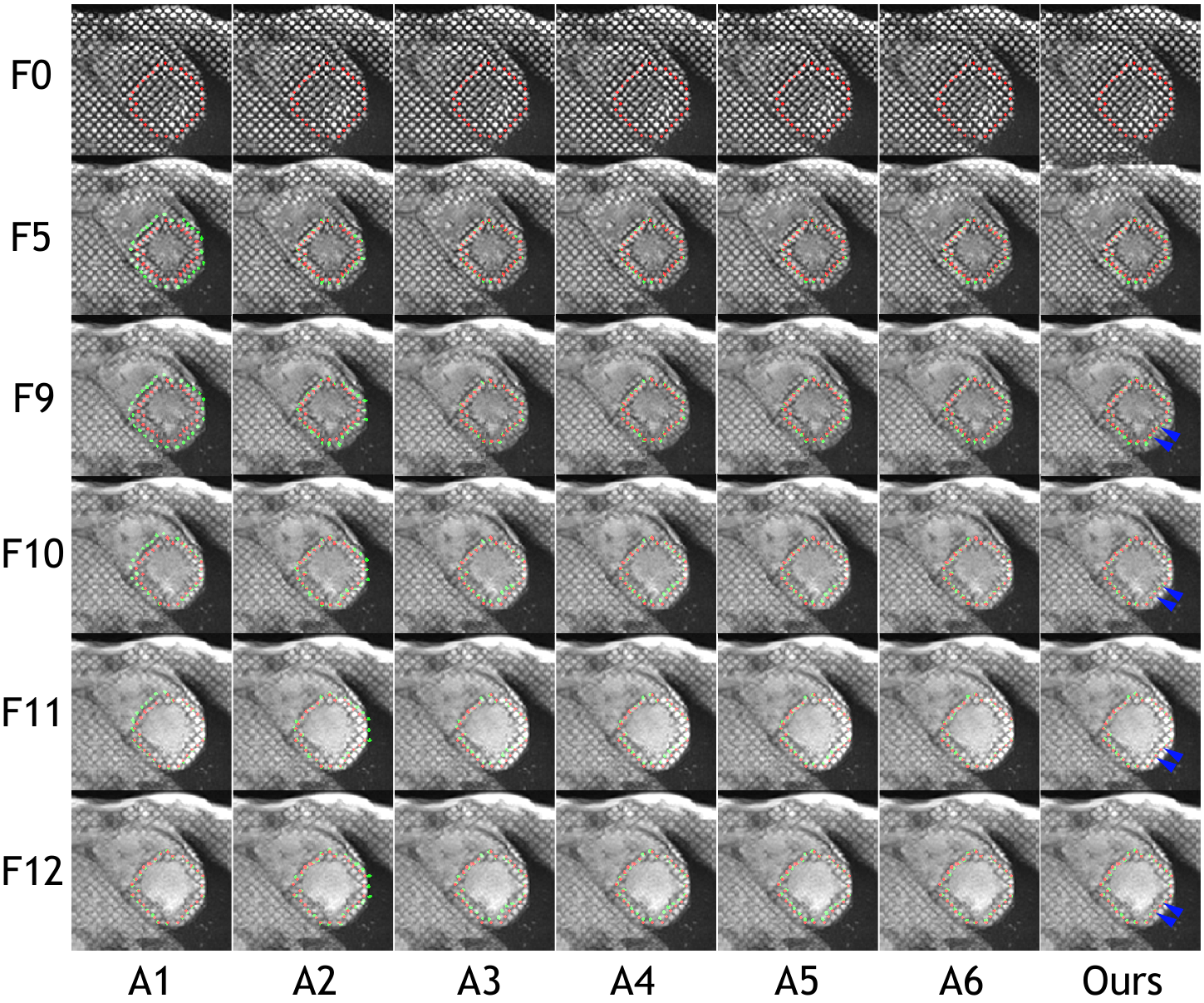}
\end{center}
   \caption{Ablation study results on an image sequence of 19 t-MRI frames. Red is ground truth, green is prediction.}
\label{fig9}
\end{figure}

\begin{figure}[t]
\begin{center}
\includegraphics[width=0.5\linewidth]{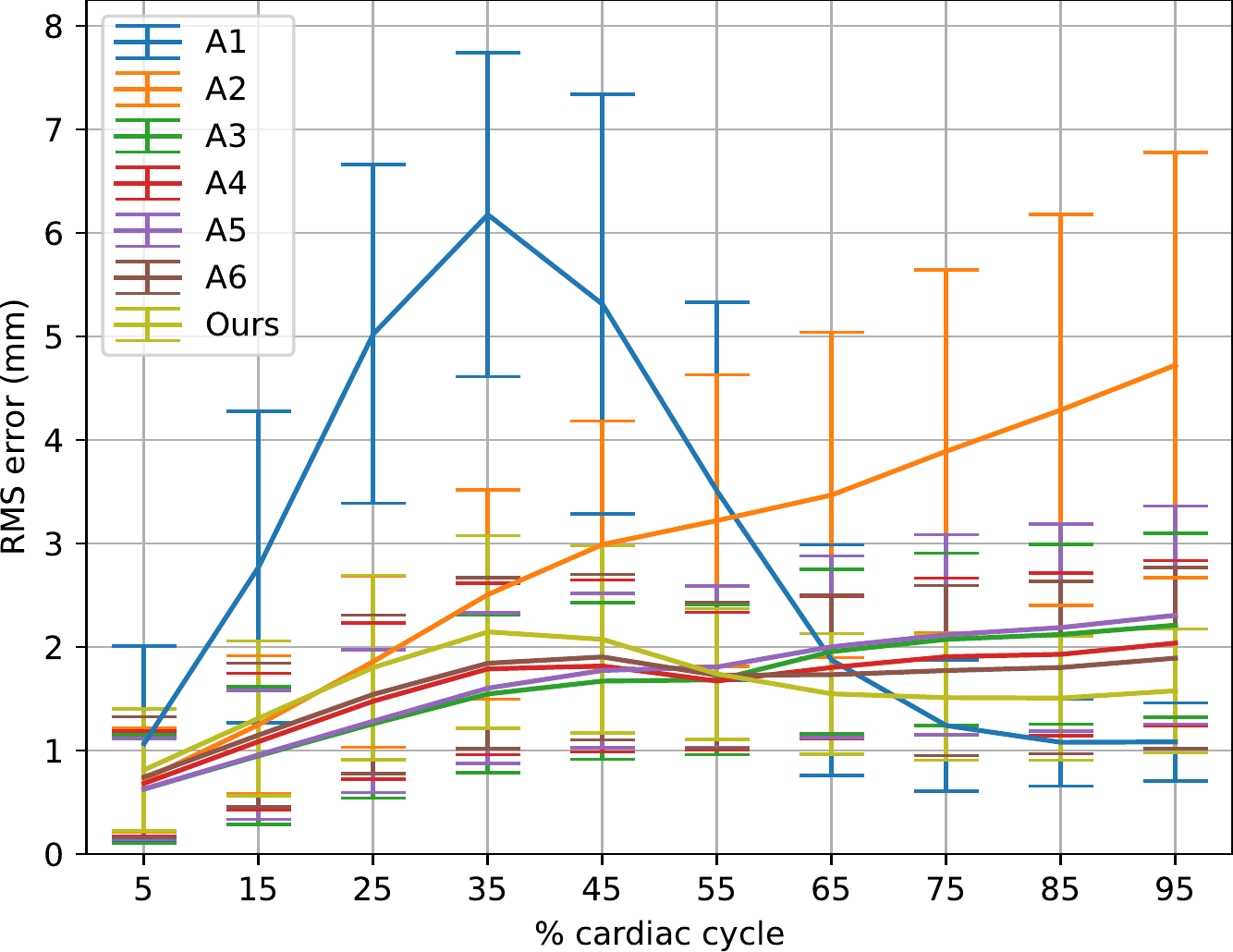}
\end{center}
   \caption{Mean and standard deviation of the RMS error during the entire cardiac cycle for the ablation study models and ours.}
\label{fig10}
\end{figure}

\subsubsection{Running Time Analysis}
In Table~\ref{table1}, we report the average inference time for motion tacking on a full t-MRI image sequence by using an Intel Xeon CPU and an NVIDIA Quadro RTX 8000 GPU for different tracking methods. While the unsupervised deep learning-based methods utilize both CPU and GPU during inference, conventional methods (HARP and OF-TV) only use the CPU.
It can be noted that the learning-based method is much faster than the conventional iteration-based method. Our method can complete the inference of the full sequence in one second. In this way, we can expect very fast and accurate regional myocardial movement tracking on t-MRI images that can be used in future clinical practice.

\section{Conclusions}
In this work, we proposed a novel bi-directional unsupervised diffeomorphic registration network to track regional myocardium motion on t-MRI images. We decomposed the Lagrangian motion tracking into a sequence of INF motion tracking, and used global constraints to correct unreasonable INF motion estimation. Experimental results on the clinical t-MRI dataset verified the effectiveness and efficiency of the proposed method. 

{\small
\bibliographystyle{ieee_fullname}
\bibliography{egbib}

\begin{thebibliography}{10}\itemsep=-1pt

\bibitem{amzulescu2019myocardial}
Mihaela~Silvia Amzulescu, M De~Craene, H Langet, Agnes Pasquet, David
  Vancraeynest, Anne-Catherine Pouleur, Jean-Louis Vanoverschelde, and BL
  Gerber.
\newblock Myocardial strain imaging: review of general principles, validation,
  and sources of discrepancies.
\newblock {\em European Heart Journal-Cardiovascular Imaging}, 20(6):605--619,
  2019.

\bibitem{arsigny2006log}
Vincent Arsigny, Olivier Commowick, Xavier Pennec, and Nicholas Ayache.
\newblock A log-euclidean framework for statistics on diffeomorphisms.
\newblock In {\em International Conference on Medical Image Computing and
  Computer-Assisted Intervention}, pages 924--931. Springer, 2006.

\bibitem{ashburner2007fast}
John Ashburner.
\newblock A fast diffeomorphic image registration algorithm.
\newblock {\em Neuroimage}, 38(1):95--113, 2007.

\bibitem{avants2011reproducible}
Brian~B Avants, Nicholas~J Tustison, Gang Song, Philip~A Cook, Arno Klein, and
  James~C Gee.
\newblock A reproducible evaluation of ants similarity metric performance in
  brain image registration.
\newblock {\em Neuroimage}, 54(3):2033--2044, 2011.

\bibitem{axel1989mr}
Leon Axel and Lawrence Dougherty.
\newblock Mr imaging of motion with spatial modulation of magnetization.
\newblock {\em Radiology}, 171(3):841--845, 1989.

\bibitem{balakrishnan2018unsupervised}
Guha Balakrishnan, Amy Zhao, Mert~R Sabuncu, John Guttag, and Adrian~V Dalca.
\newblock An unsupervised learning model for deformable medical image
  registration.
\newblock In {\em Proceedings of the IEEE conference on computer vision and
  pattern recognition}, pages 9252--9260, 2018.

\bibitem{brox2004high}
Thomas Brox, Andr{\'e}s Bruhn, Nils Papenberg, and Joachim Weickert.
\newblock High accuracy optical flow estimation based on a theory for warping.
\newblock In {\em European conference on computer vision}, pages 25--36.
  Springer, 2004.

\bibitem{brox2010large}
Thomas Brox and Jitendra Malik.
\newblock Large displacement optical flow: descriptor matching in variational
  motion estimation.
\newblock {\em IEEE transactions on pattern analysis and machine intelligence},
  33(3):500--513, 2010.

\bibitem{cao2017deformable}
Xiaohuan Cao, Jianhua Yang, Jun Zhang, Dong Nie, Minjeong Kim, Qian Wang, and
  Dinggang Shen.
\newblock Deformable image registration based on similarity-steered cnn
  regression.
\newblock In {\em International Conference on Medical Image Computing and
  Computer-Assisted Intervention}, pages 300--308. Springer, 2017.

\bibitem{cao2018deformable}
Xiaohuan Cao, Jianhua Yang, Jun Zhang, Qian Wang, Pew-Thian Yap, and Dinggang
  Shen.
\newblock Deformable image registration using a cue-aware deep regression
  network.
\newblock {\em IEEE Transactions on Biomedical Engineering}, 65(9):1900--1911,
  2018.

\bibitem{carranza2010motion}
Noemi Carranza-Herrezuelo, Ana Bajo, Filip Sroubek, Cristina Santamarta,
  Gabriel Crist{\'o}bal, Andr{\'e}s Santos, and Mar{\'\i}a~J Ledesma-Carbayo.
\newblock Motion estimation of tagged cardiac magnetic resonance images using
  variational techniques.
\newblock {\em Computerized Medical Imaging and Graphics}, 34(6):514--522,
  2010.

\bibitem{chandrashekara2004analysis}
Raghavendra Chandrashekara, Raad~H Mohiaddin, and Daniel Rueckert.
\newblock Analysis of 3-d myocardial motion in tagged mr images using nonrigid
  image registration.
\newblock {\em IEEE Transactions on Medical Imaging}, 23(10):1245--1250, 2004.

\bibitem{chen2009automated}
Ting Chen, Xiaoxu Wang, Sohae Chung, Dimitris Metaxas, and Leon Axel.
\newblock Automated 3d motion tracking using gabor filter bank, robust point
  matching, and deformable models.
\newblock {\em IEEE Transactions on Medical Imaging}, 29(1):1--11, 2009.

\bibitem{dalca2018unsupervised}
Adrian~V Dalca, Guha Balakrishnan, John Guttag, and Mert~R Sabuncu.
\newblock Unsupervised learning for fast probabilistic diffeomorphic
  registration.
\newblock In {\em International Conference on Medical Image Computing and
  Computer-Assisted Intervention}, pages 729--738. Springer, 2018.

\bibitem{de2017end}
Bob~D de Vos, Floris~F Berendsen, Max~A Viergever, Marius Staring, and Ivana
  I{\v{s}}gum.
\newblock End-to-end unsupervised deformable image registration with a
  convolutional neural network.
\newblock In {\em Deep Learning in Medical Image Analysis and Multimodal
  Learning for Clinical Decision Support}, pages 204--212. Springer, 2017.

\bibitem{dosovitskiy2015flownet}
Alexey Dosovitskiy, Philipp Fischer, Eddy Ilg, Philip Hausser, Caner Hazirbas,
  Vladimir Golkov, Patrick Van Der~Smagt, Daniel Cremers, and Thomas Brox.
\newblock Flownet: Learning optical flow with convolutional networks.
\newblock In {\em Proceedings of the IEEE international conference on computer
  vision}, pages 2758--2766, 2015.

\bibitem{eldeeb2016accurate}
Safaa~M ElDeeb and Ahmed~S Fahmy.
\newblock Accurate harmonic phase tracking of tagged mri using locally-uniform
  myocardium displacement constraint.
\newblock {\em Medical Engineering \& Physics}, 38(11):1305--1313, 2016.

\bibitem{horn1981determining}
Berthold~KP Horn and Brian~G Schunck.
\newblock Determining optical flow.
\newblock In {\em Techniques and Applications of Image Understanding}, volume
  281, pages 319--331. International Society for Optics and Photonics, 1981.

\bibitem{hur2019iterative}
Junhwa Hur and Stefan Roth.
\newblock Iterative residual refinement for joint optical flow and occlusion
  estimation.
\newblock In {\em Proceedings of the IEEE Conference on Computer Vision and
  Pattern Recognition}, pages 5754--5763, 2019.

\bibitem{ilg2017flownet}
Eddy Ilg, Nikolaus Mayer, Tonmoy Saikia, Margret Keuper, Alexey Dosovitskiy,
  and Thomas Brox.
\newblock Flownet 2.0: Evolution of optical flow estimation with deep networks.
\newblock In {\em Proceedings of the IEEE conference on computer vision and
  pattern recognition}, pages 2462--2470, 2017.

\bibitem{kingma2013auto}
Diederik~P Kingma and Max Welling.
\newblock Auto-encoding variational bayes.
\newblock {\em arXiv preprint arXiv:1312.6114}, 2013.

\bibitem{krebs2019learning}
Julian Krebs, Herv{\'e} Delingette, Boris Mailh{\'e}, Nicholas Ayache, and
  Tommaso Mansi.
\newblock Learning a probabilistic model for diffeomorphic registration.
\newblock {\em IEEE transactions on medical imaging}, 38(9):2165--2176, 2019.

\bibitem{krebs2017robust}
Julian Krebs, Tommaso Mansi, Herv{\'e} Delingette, Li Zhang, Florin~C Ghesu,
  Shun Miao, Andreas~K Maier, Nicholas Ayache, Rui Liao, and Ali Kamen.
\newblock Robust non-rigid registration through agent-based action learning.
\newblock In {\em International Conference on Medical Image Computing and
  Computer-Assisted Intervention}, pages 344--352. Springer, 2017.

\bibitem{lai2019bridging}
Hsueh-Ying Lai, Yi-Hsuan Tsai, and Wei-Chen Chiu.
\newblock Bridging stereo matching and optical flow via spatiotemporal
  correspondence.
\newblock In {\em Proceedings of the IEEE Conference on Computer Vision and
  Pattern Recognition}, pages 1890--1899, 2019.

\bibitem{ledesma2008unsupervised}
Maria~J Ledesma-Carbayo, J~Andrew Derbyshire, Smita Sampath, Andr{\'e}s Santos,
  Manuel Desco, and Elliot~R McVeigh.
\newblock Unsupervised estimation of myocardial displacement from tagged mr
  sequences using nonrigid registration.
\newblock {\em Magnetic Resonance in Medicine: An Official Journal of the
  International Society for Magnetic Resonance in Medicine}, 59(1):181--189,
  2008.

\bibitem{liu2019selflow}
Pengpeng Liu, Michael Lyu, Irwin King, and Jia Xu.
\newblock Selflow: Self-supervised learning of optical flow.
\newblock In {\em Proceedings of the IEEE Conference on Computer Vision and
  Pattern Recognition}, pages 4571--4580, 2019.

\bibitem{liu2011incompressible}
Xiaofeng Liu, Khaled~Z Abd-Elmoniem, Maureen Stone, Emi~Z Murano, Jiachen Zhuo,
  Rao~P Gullapalli, and Jerry~L Prince.
\newblock Incompressible deformation estimation algorithm (idea) from tagged mr
  images.
\newblock {\em IEEE transactions on medical imaging}, 31(2):326--340, 2011.

\bibitem{liu2010shortest}
Xiaofeng Liu and Jerry~L Prince.
\newblock Shortest path refinement for motion estimation from tagged mr images.
\newblock {\em IEEE Transactions on Medical Imaging}, 29(8):1560--1572, 2010.

\bibitem{lorenzi2013lcc}
Marco Lorenzi, Nicholas Ayache, Giovanni~B Frisoni, Xavier Pennec, Alzheimer's
  Disease Neuroimaging~Initiative (ADNI, et~al.
\newblock Lcc-demons: a robust and accurate symmetric diffeomorphic
  registration algorithm.
\newblock {\em NeuroImage}, 81:470--483, 2013.

\bibitem{mcleod2011incompressible}
Kristin McLeod, Adityo Prakosa, Tommaso Mansi, Maxime Sermesant, and Xavier
  Pennec.
\newblock An incompressible log-domain demons algorithm for tracking heart
  tissue.
\newblock In {\em International Workshop on Statistical Atlases and
  Computational Models of the Heart}, pages 55--67. Springer, 2011.

\bibitem{meister2017unflow}
Simon Meister, Junhwa Hur, and Stefan Roth.
\newblock Unflow: Unsupervised learning of optical flow with a bidirectional
  census loss.
\newblock {\em arXiv preprint arXiv:1711.07837}, 2017.

\bibitem{memin1998dense}
Etienne M{\'e}min and Patrick P{\'e}rez.
\newblock Dense estimation and object-based segmentation of the optical flow
  with robust techniques.
\newblock {\em IEEE Transactions on Image Processing}, 7(5):703--719, 1998.

\bibitem{mok2020fast}
Tony~CW Mok and Albert Chung.
\newblock Fast symmetric diffeomorphic image registration with convolutional
  neural networks.
\newblock In {\em Proceedings of the IEEE/CVF Conference on Computer Vision and
  Pattern Recognition}, pages 4644--4653, 2020.

\bibitem{morais2013cardiac}
Pedro Morais, Brecht Heyde, Daniel Barbosa, Sandro Queir{\'o}s, Piet Claus, and
  Jan D’hooge.
\newblock Cardiac motion and deformation estimation from tagged mri sequences
  using a temporal coherent image registration framework.
\newblock In {\em International Conference on Functional Imaging and Modeling
  of the Heart}, pages 316--324. Springer, 2013.

\bibitem{morales2019implementation}
Manuel~A Morales, David Izquierdo-Garcia, Iman Aganj, Jayashree
  Kalpathy-Cramer, Bruce~R Rosen, and Ciprian Catana.
\newblock Implementation and validation of a three-dimensional cardiac motion
  estimation network.
\newblock {\em Radiology: Artificial Intelligence}, 1(4):e180080, 2019.

\bibitem{niethammer2019metric}
Marc Niethammer, Roland Kwitt, and Francois-Xavier Vialard.
\newblock Metric learning for image registration.
\newblock In {\em Proceedings of the IEEE Conference on Computer Vision and
  Pattern Recognition}, pages 8463--8472, 2019.

\bibitem{osman1999cardiac}
Nael~F Osman, William~S Kerwin, Elliot~R McVeigh, and Jerry~L Prince.
\newblock Cardiac motion tracking using cine harmonic phase (harp) magnetic
  resonance imaging.
\newblock {\em Magnetic Resonance in Medicine: An Official Journal of the
  International Society for Magnetic Resonance in Medicine}, 42(6):1048--1060,
  1999.

\bibitem{osman2000imaging}
Nael~F Osman, Elliot~R McVeigh, and Jerry~L Prince.
\newblock Imaging heart motion using harmonic phase mri.
\newblock {\em IEEE transactions on medical imaging}, 19(3):186--202, 2000.

\bibitem{qian2011identifying}
Zhen Qian, Qingshan Liu, Dimitris~N Metaxas, and Leon Axel.
\newblock Identifying regional cardiac abnormalities from myocardial strains
  using nontracking-based strain estimation and spatio-temporal tensor
  analysis.
\newblock {\em IEEE Transactions on Medical Imaging}, 30(12):2017--2029, 2011.

\bibitem{qin2018joint}
Chen Qin, Wenjia Bai, Jo Schlemper, Steffen~E Petersen, Stefan~K Piechnik,
  Stefan Neubauer, and Daniel Rueckert.
\newblock Joint learning of motion estimation and segmentation for cardiac mr
  image sequences.
\newblock In {\em International Conference on Medical Image Computing and
  Computer-Assisted Intervention}, pages 472--480. Springer, 2018.

\bibitem{ranjan2019competitive}
Anurag Ranjan, Varun Jampani, Lukas Balles, Kihwan Kim, Deqing Sun, Jonas
  Wulff, and Michael~J Black.
\newblock Competitive collaboration: Joint unsupervised learning of depth,
  camera motion, optical flow and motion segmentation.
\newblock In {\em Proceedings of the IEEE conference on computer vision and
  pattern recognition}, pages 12240--12249, 2019.

\bibitem{rohe2017svf}
Marc-Michel Roh{\'e}, Manasi Datar, Tobias Heimann, Maxime Sermesant, and
  Xavier Pennec.
\newblock Svf-net: Learning deformable image registration using shape matching.
\newblock In {\em International conference on medical image computing and
  computer-assisted intervention}, pages 266--274. Springer, 2017.

\bibitem{rougon2005non}
Nicolas Rougon, Caroline Petitjean, Fran{\c{c}}oise Pr{\^e}teux, Philippe
  Cluzel, and Philippe Grenier.
\newblock A non-rigid registration approach for quantifying myocardial
  contraction in tagged mri using generalized information measures.
\newblock {\em Medical Image Analysis}, 9(4):353--375, 2005.

\bibitem{shen2019networks}
Zhengyang Shen, Xu Han, Zhenlin Xu, and Marc Niethammer.
\newblock Networks for joint affine and non-parametric image registration.
\newblock In {\em Proceedings of the IEEE Conference on Computer Vision and
  Pattern Recognition}, pages 4224--4233, 2019.

\bibitem{shen2019region}
Zhengyang Shen, Fran{\c{c}}ois-Xavier Vialard, and Marc Niethammer.
\newblock Region-specific diffeomorphic metric mapping.
\newblock In {\em Advances in Neural Information Processing Systems}, pages
  1098--1108, 2019.

\bibitem{shi2012comprehensive}
Wenzhe Shi, Xiahai Zhuang, Haiyan Wang, Simon Duckett, Duy~VN Luong, Catalina
  Tobon-Gomez, KaiPin Tung, Philip~J Edwards, Kawal~S Rhode, Reza~S Razavi,
  et~al.
\newblock A comprehensive cardiac motion estimation framework using both
  untagged and 3-d tagged mr images based on nonrigid registration.
\newblock {\em IEEE transactions on medical imaging}, 31(6):1263--1275, 2012.

\bibitem{sun2018pwc}
Deqing Sun, Xiaodong Yang, Ming-Yu Liu, and Jan Kautz.
\newblock Pwc-net: Cnns for optical flow using pyramid, warping, and cost
  volume.
\newblock In {\em Proceedings of the IEEE conference on computer vision and
  pattern recognition}, pages 8934--8943, 2018.

\bibitem{teed2020raft}
Zachary Teed and Jia Deng.
\newblock Raft: Recurrent all-pairs field transforms for optical flow.
\newblock {\em arXiv preprint arXiv:2003.12039}, 2020.

\bibitem{wang2019gradient}
Liang Wang, Patrick Clarysse, Zhengjun Liu, Bin Gao, Wanyu Liu, Pierre
  Croisille, and Philippe Delachartre.
\newblock A gradient-based optical-flow cardiac motion estimation method for
  cine and tagged mr images.
\newblock {\em Medical image analysis}, 57:136--148, 2019.

\bibitem{wang2008meshless}
Xiaoxu Wang, Dimitis Metaxas, Ting Chen, and Leon Axel.
\newblock Meshless deformable models for lv motion analysis.
\newblock In {\em 2008 IEEE Conference on Computer Vision and Pattern
  Recognition}, pages 1--8. IEEE, 2008.

\bibitem{wang2019unos}
Yang Wang, Peng Wang, Zhenheng Yang, Chenxu Luo, Yi Yang, and Wei Xu.
\newblock Unos: Unified unsupervised optical-flow and stereo-depth estimation
  by watching videos.
\newblock In {\em Proceedings of the IEEE Conference on Computer Vision and
  Pattern Recognition}, pages 8071--8081, 2019.

\bibitem{wedel2009structure}
Andreas Wedel, Daniel Cremers, Thomas Pock, and Horst Bischof.
\newblock Structure-and motion-adaptive regularization for high accuracy optic
  flow.
\newblock In {\em 2009 IEEE 12th International Conference on Computer Vision},
  pages 1663--1668. IEEE, 2009.

\bibitem{yin2018geonet}
Zhichao Yin and Jianping Shi.
\newblock Geonet: Unsupervised learning of dense depth, optical flow and camera
  pose.
\newblock In {\em Proceedings of the IEEE Conference on Computer Vision and
  Pattern Recognition}, pages 1983--1992, 2018.

\bibitem{yu2020motion}
Hanchao Yu, Xiao Chen, Humphrey Shi, Terrence Chen, Thomas~S Huang, and Shanhui
  Sun.
\newblock Motion pyramid networks for accurate and efficient cardiac motion
  estimation.
\newblock In {\em Medical Image Computing and Computer Assisted
  Intervention--MICCAI 2020: 23rd International Conference, Lima, Peru, October
  4--8, 2020, Proceedings, Part VI 23}, pages 436--446. Springer, 2020.

\bibitem{yu2020foal}
Hanchao Yu, Shanhui Sun, Haichao Yu, Xiao Chen, Honghui Shi, Thomas~S Huang,
  and Terrence Chen.
\newblock Foal: Fast online adaptive learning for cardiac motion estimation.
\newblock In {\em Proceedings of the IEEE/CVF Conference on Computer Vision and
  Pattern Recognition}, pages 4313--4323, 2020.

\bibitem{zhao2019recursive}
Shengyu Zhao, Yue Dong, Eric~I Chang, Yan Xu, et~al.
\newblock Recursive cascaded networks for unsupervised medical image
  registration.
\newblock In {\em Proceedings of the IEEE International Conference on Computer
  Vision}, pages 10600--10610, 2019.

\bibitem{zheng2019explainable}
Qiao Zheng, Herv{\'e} Delingette, and Nicholas Ayache.
\newblock Explainable cardiac pathology classification on cine mri with motion
  characterization by semi-supervised learning of apparent flow.
\newblock {\em Medical image analysis}, 56:80--95, 2019.

\end{thebibliography}
}

\clearpage
\section{Supplementary Material}
\begin{figure}[t]
\begin{center}
\includegraphics[width=1.0\linewidth]{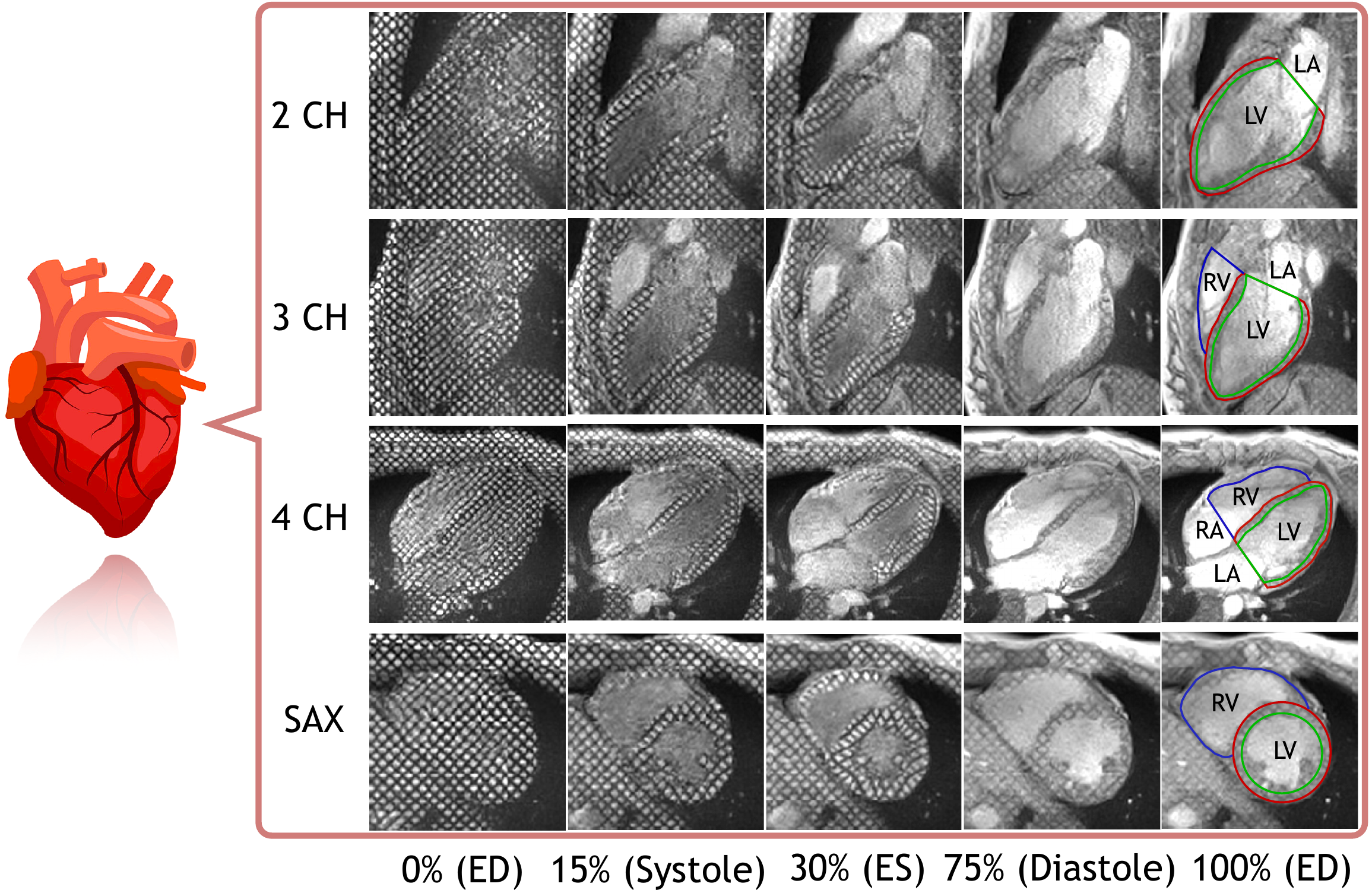}
\end{center}
   \caption{Typical scan views (2-, 3-, 4-chamber views and short-axis view) of cardiac tagging MRI. Number under the figure means percentage of one cardiac cycle. Red and green contours depict the epi- and endo-cardial borders of left ventricle (LV) myocardium (MYO) wall. Bright area within the green contour is the LV blood pool. Blue contour depicts the right ventricle (RV). LA: left atrium. RA: right atrium.}
\label{figs1}
\end{figure}

\subsection{Challenges for T-MRI Motion Tracking}
Tags are physical properties of the tissue which will deform with the heart, as it contracts and relaxes during a cardiac cycle. Tracking the deformation of tags can help us retrieve a 2D displacement field in the imaging plane and reconstruct local motion of the myocardium. Challenges for motion tracking on cardiac tagging magnetic resonance imaging (t-MRI) images can include the following. 

(1) Image appearance changes a lot even within a 2D sequence. One can observe in Fig.~\ref{figs1} that, at the beginning of the heart cycle, which is the end diastole (ED) phase, tag grids cover the imaging plane except for the background lung cavity. As the heart contracts to the end systole (ES) phase, approximately $30\%$ of a cycle, untagged blood replace tagged blood, leaving a brighter untagged blood pool. After ES phase, as the heart relaxes towards the ED phase ($100\%$ of a cycle), tag signal in the myocardium fades because of T1 relaxation of the perturbed magnetization vectors. So the dark tags get brighter and brighter in the later frames. This can pose a big challenge for the tag deformation estimation. 

(2) Different frames can have a very different image appearance, related to the changing myocardium shape. 

(3) t-MRI images have relatively low temporal resolution, reaching at the best $20\sim30$ frames in a cycle. Motion in between consecutive two frames could be large, especially during the rapid early filling phase. 

(4) t-MRI is 2D imaging, and through-plane motion of the heart through the fixed imaging plane could make tags disappear at one frame and reappear at some other frame within a sequence. Non identification of such tags will incur motion-tracking errors. 

(5) Due to imaging condition changes in the imaging process, such as magnetic field drift and patients' respiratory motion by unsuccessful breath holding, other artifacts and noise can degrade image quality.

\subsection{Derivation of KL Loss}
Detailed derivation of the KL loss is as following:
\begin{equation}
\begin{aligned}
    &\mathcal{KL}[q_{\boldsymbol{\psi} }(\boldsymbol{z}|\boldsymbol{y};\boldsymbol{x})||p(\boldsymbol{z}|\boldsymbol{y};\boldsymbol{x})] 
    \\&=\mathbb{E}_{q}\left [ log\frac{q_{\boldsymbol{\psi} }(\boldsymbol{z}|\boldsymbol{y};\boldsymbol{x})}{p(\boldsymbol{z}|\boldsymbol{y};\boldsymbol{x})} \right ]
    \\&=\mathbb{E}_{q}\left [ log\frac{q_{\boldsymbol{\psi} }(\boldsymbol{z}|\boldsymbol{y};\boldsymbol{x})p(\boldsymbol{y};\boldsymbol{x})}{p(\boldsymbol{z}|\boldsymbol{y};\boldsymbol{x})p(\boldsymbol{y};\boldsymbol{x})} \right ]
    \\&=\mathbb{E}_{q}\left [ log\frac{q_{\boldsymbol{\psi} }(\boldsymbol{z}|\boldsymbol{y};\boldsymbol{x})p(\boldsymbol{y};\boldsymbol{x})}{p(\boldsymbol{z},\boldsymbol{y};\boldsymbol{x})} \right ]
    \\&=\mathbb{E}_{q}\left [ log\frac{q_{\boldsymbol{\psi} }(\boldsymbol{z}|\boldsymbol{y};\boldsymbol{x})p(\boldsymbol{y};\boldsymbol{x})}{p(\boldsymbol{y}|\boldsymbol{z};\boldsymbol{x})p(\boldsymbol{x}|\boldsymbol{z})p(\boldsymbol{z})} \right ]
    \\&=\mathbb{E}_{q}\left [ log\frac{q_{\boldsymbol{\psi} }(\boldsymbol{z}|\boldsymbol{y};\boldsymbol{x})p(\boldsymbol{y};\boldsymbol{x})}{p(\boldsymbol{y}|\boldsymbol{z};\boldsymbol{x})p(\boldsymbol{x})p(\boldsymbol{z})} \right ]
    \\&=\mathbb{E}_{q}\left [ log\frac{q_{\boldsymbol{\psi} }(\boldsymbol{z}|\boldsymbol{y};\boldsymbol{x})}{p(\boldsymbol{z})} \right ] - \mathbb{E}_{q}\left [ log\ p(\boldsymbol{y}|\boldsymbol{z};\boldsymbol{x}) \right ] + log\ p(\boldsymbol{y}|\boldsymbol{x})
    \\&=\mathcal{KL}[q_{\boldsymbol{\psi} }(\boldsymbol{z}|\boldsymbol{y};\boldsymbol{x})||p(\boldsymbol{z})]- \mathbb{E}_{q}\left [log\ p(\boldsymbol{y}|\boldsymbol{z};\boldsymbol{x}) \right ] + const.
\end{aligned}
\label{eqs1}
\end{equation}

For the first term, we have
\begin{equation}
\begin{aligned}
    &\mathcal{KL}[q_{\boldsymbol{\psi} }(\boldsymbol{z}|\boldsymbol{y};\boldsymbol{x})||p(\boldsymbol{z})]
    \\&=\mathcal{KL}[\mathcal N(\boldsymbol{z};\boldsymbol{\mu}_{z|x,y},\boldsymbol{\Sigma }_{z|x,y})|| \mathcal N(\boldsymbol{z};\boldsymbol{0},\boldsymbol{\Sigma} _{z})]
    \\&=\frac{1}{2}[log\frac{|\boldsymbol{\Sigma} _{z}|}{|\boldsymbol{\Sigma }_{z|x,y}|}-n+tr(\boldsymbol{\Sigma} _{z}^{-1}\boldsymbol{\Sigma }_{z|x,y}) \\&+\boldsymbol{\mu}_{z|x,y}^{T}\boldsymbol{\Sigma }_{z}^{-1}\boldsymbol{\mu}_{z|x,y}],
\end{aligned}
\label{eqs2}
\end{equation}
where $n$ is the total number of the variables in $p(\boldsymbol{z})$.

According to our setting, $\boldsymbol{\Lambda }_{z}=\boldsymbol{\Sigma }_{z}^{-1}= \lambda\boldsymbol{L}$~\cite{dalca2018unsupervised}, where $\boldsymbol{L}=\boldsymbol{D}-\boldsymbol{A}$ is the Laplacian of a neighborhood graph defined on the pixel grid, $\boldsymbol{D}$ is the graph degree matrix, $\boldsymbol{A}$ is a pixel neighborhood adjacency matrix. Therefore, $log|\boldsymbol{\Sigma} _{z}|$ is constant. Since $\boldsymbol{\Sigma }_{z|x,y}$ is set to be diagonal, $log|\boldsymbol{\Sigma }_{z|x,y}|=tr \ log\boldsymbol{\Sigma }_{z|x,y}$. And $tr(\boldsymbol{\Sigma} _{z}^{-1}\boldsymbol{\Sigma }_{z|x,y})=tr(\lambda(\boldsymbol{D}-\boldsymbol{A})\boldsymbol{\Sigma }_{z|x,y})=tr(\lambda\boldsymbol{D}\boldsymbol{\Sigma }_{z|x,y})$. So we can get
\begin{equation}
\begin{aligned}
    &\mathcal{KL}[q_{\boldsymbol{\psi} }(\boldsymbol{z}|\boldsymbol{y};\boldsymbol{x})||p(\boldsymbol{z})]
    \\&=\frac{1}{2}[tr(\lambda\boldsymbol{D}\boldsymbol{\Sigma }_{z|x,y}-log\boldsymbol{\Sigma }_{z|x,y})+\boldsymbol{\mu}_{z|x,y}^{T}\boldsymbol{\Lambda }_{z}\boldsymbol{\mu}_{z|x,y}]
    \\&+ const.
\end{aligned}
\label{eqs3}
\end{equation}

For the second term, if we model $p(\boldsymbol{y}|\boldsymbol{z};\boldsymbol{x})$ as a Gaussian, we can get
\begin{equation}
\begin{aligned}
    &-\mathbb{E}_{q}\left [log\ p(\boldsymbol{y}|\boldsymbol{z};\boldsymbol{x}) \right ]
    \\&=-\mathbb{E}_{q}[log\ \mathcal N(\boldsymbol{y};\boldsymbol{x}\circ\boldsymbol{\phi},\sigma^{2}\mathbb{I})]
    \\&=\frac{1}{2}\mathbb{E}_{q}[log\ (2\pi\sigma ^{2})^{n}+\frac{1}{\sigma ^{2}}\left \| \boldsymbol{y}-\boldsymbol{x}\circ\boldsymbol{\phi} \right \|_{2}^{2}]
    \\&=\frac{1}{2\sigma ^{2}}\mathbb{E}_{q}[\left \| \boldsymbol{y}-\boldsymbol{x}\circ\boldsymbol{\phi} \right \|_{2}^{2}]+const,
\end{aligned}
\label{eqs4}
\end{equation}
where the term $\left \| \boldsymbol{y}-\boldsymbol{x}\circ\boldsymbol{\phi} \right \|_{2}^{2}$ corresponds to the sum-of-squared difference (SSD) metric.

If we model $p(\boldsymbol{y}|\boldsymbol{z};\boldsymbol{x})$ as a Boltzmann distribution, we can get
\begin{equation}
\begin{aligned}
    &-\mathbb{E}_{q}\left [log\ p(\boldsymbol{y}|\boldsymbol{z};\boldsymbol{x}) \right ]
    \\&\propto -\mathbb{E}_{q}[log\ exp(-\gamma NCC(\boldsymbol{y}, \boldsymbol{x}\circ\boldsymbol{\phi}))]
    \\&=\gamma \mathbb{E}_{q}[NCC(\boldsymbol{y}, \boldsymbol{x}\circ\boldsymbol{\phi})],
\end{aligned}
\label{eqs5}
\end{equation}
where $\gamma$ is a negative scalar hyperparameter, NCC is the normalized local cross correlation metric.

We can approximate the expectation $\mathbb{E}_{q}$ with $K$ samples $\boldsymbol{z}_{k}\sim q_{\boldsymbol{z}}$, so we get
\begin{equation}
\begin{aligned}
    -\mathbb{E}_{q}\left [log\ p(\boldsymbol{y}|\boldsymbol{z};\boldsymbol{x}) \right ]=\frac{\gamma}{K}\sum_{k}[NCC(\boldsymbol{y}, \boldsymbol{x}\circ\boldsymbol{\phi}_{k})].
\end{aligned}
\label{eqs6}
\end{equation}

Note that by Eq.~(\ref{eqs1}), we can get:
\begin{equation}
\begin{aligned}
    &log\ p(\boldsymbol{y}|\boldsymbol{x}) = \mathcal{KL}[q_{\boldsymbol{\psi} }(\boldsymbol{z}|\boldsymbol{y};\boldsymbol{x})||p(\boldsymbol{z}|\boldsymbol{y};\boldsymbol{x})] 
    \\&-(\mathcal{KL}[q_{\boldsymbol{\psi} }(\boldsymbol{z}|\boldsymbol{y};\boldsymbol{x})||p(\boldsymbol{z})]-\mathbb{E}_{q}\left [log\ p(\boldsymbol{y}|\boldsymbol{z};\boldsymbol{x}) \right ])
    \\&=\mathcal{KL}[q_{\boldsymbol{\psi} }(\boldsymbol{z}|\boldsymbol{y};\boldsymbol{x})||p(\boldsymbol{z}|\boldsymbol{y};\boldsymbol{x})]+ELBO
    \\&\geq ELBO,
\end{aligned}
\label{eqelbo}
\end{equation}
where 
\begin{equation}
    ELBO=-(\mathcal{KL}[q_{\boldsymbol{\psi}}(\boldsymbol{z}|\boldsymbol{y};\boldsymbol{x})||p(\boldsymbol{z})]-\mathbb{E}_{q}\left [log\ p(\boldsymbol{y}|\boldsymbol{z};\boldsymbol{x}) \right ]).
\end{equation}
Thus, maximizing the ELBO of the log marginalized likelihood $log\ p(\boldsymbol{y}|\boldsymbol{x})$ in Eq.~(\ref{eqelbo}) is equivalent to minimizing $\mathcal{KL}[q_{\boldsymbol{\psi} }(\boldsymbol{z}|\boldsymbol{y};\boldsymbol{x})||p(\boldsymbol{z}|\boldsymbol{y};\boldsymbol{x})]$ in Eq.~(\ref{eqs1}).

\subsection{Backward Registration}
With the SVF representation, we can also compute an inverse motion field $\boldsymbol{\phi}^{-1}$ by inputting $-\boldsymbol{z}$ into the SS layer: $\boldsymbol{\phi}^{-1}=SS(-\boldsymbol{z})$. Thus we can warp image $\boldsymbol{y}$ backward towards image $\boldsymbol{x}$ and get the observation distribution of warped image, $\boldsymbol{y}\circ\boldsymbol{\phi}^{-1}$, which is also modeled by a Boltzmann distribution:
\begin{equation}
     p(\boldsymbol{x}|\boldsymbol{z};\boldsymbol{y})\sim exp(-\gamma NCC(\boldsymbol{x}, \boldsymbol{y}\circ\boldsymbol{\phi}^{-1})),
\end{equation}
where $\boldsymbol{x}$ denotes the observation of warped image $\boldsymbol{y}$. We call the process of warping image $\boldsymbol{y}$ towards $\boldsymbol{x}$ the backward registration. We minimize the KL divergence between $q_{\boldsymbol{\psi} }(\boldsymbol{z}|\boldsymbol{x};\boldsymbol{y})$ and $p(\boldsymbol{z}|\boldsymbol{x};\boldsymbol{y})$, which leads to maximizing the ELBO of the log marginalized likelihood $log\ p(\boldsymbol{x}|\boldsymbol{y})$ as follows:
\begin{equation}
\begin{aligned}
    &\underset{\boldsymbol{\psi}}{min}\, \, \mathcal{KL}[q_{\boldsymbol{\psi} }(\boldsymbol{z}|\boldsymbol{x};\boldsymbol{y})||p(\boldsymbol{z}|\boldsymbol{x};\boldsymbol{y})]
    \\ & =\underset{\boldsymbol{\psi}}{min}\, \, \mathcal{KL}[q_{\boldsymbol{\psi} }(\boldsymbol{z}|\boldsymbol{x};\boldsymbol{y})||p(\boldsymbol{z})]
     -\mathbb{E}_{q}[log\, p(\boldsymbol{x}|\boldsymbol{z};\boldsymbol{y})] 
     \\ &+log\ p(\boldsymbol{x}|\boldsymbol{y}).
\end{aligned}
\label{eqbackward}
\end{equation}

\subsection{Algorithm 1}
 We use Algorithm~\ref{alg1} to compute Lagrangian motion field $\boldsymbol{\Phi}$ from interframe (INF) motion field $\boldsymbol{\phi}$.

\begin{algorithm}
    \caption{INF motion field recomposition as Lagrangian motion field}
    \LinesNumbered
    \label{alg1} 
    \KwIn{INF motion field: $\boldsymbol{\phi}$}
    \KwOut{Lagrangian motion field: $\boldsymbol{\Phi}$}
    
    $//$ shape: frames, channels, size x, size y\\\hspace{-0.1cm}
    $shape$$\leftarrow $shape of $\boldsymbol{\phi}$\;
    $N$$\leftarrow $$shape[0]$\;
    $C$$\leftarrow $$shape[1]$\;
    $\boldsymbol{\Phi}$$\leftarrow $$\boldsymbol{\phi}$\;
    \For{$n$ from $1$ to $N-1$ }{
        $src$$\leftarrow$$\boldsymbol{\phi}[n,::]$\;
        $flow$$\leftarrow$$\boldsymbol{\Phi}[n-1,::]$\;
        \For{$d$ from $0$ to $C-1$}{
        $//$ $\circ$ is linear interpolation 
        $\boldsymbol{\Phi}[n,d,::]$$\leftarrow $$flow[0,d,::]$\\\hspace{1.9cm}
        $+src[0,d,::]$$\circ$$flow$\;
        }
    }
\end{algorithm}

\subsection{Evaluation Metric Definitions}
Following the metric used in~\cite{chandrashekara2004analysis}, we use the root mean squared (RMS) error of distance between the centers of predicted landmark ${X}'$ and ground truth landmark $X$ for evaluation of motion tracking accuracy:
\begin{equation}
    RMS=\sqrt{\frac{1}{M}\sum_{i=0}^{M-1}\left \| X_{i}^{'} -X_{i}\right \|_{2}^{2}},
\end{equation}
where $M$ is the total number of predefined ground truth landmarks. 

In addition, we evaluate the diffeomorphic property of the predicted INF motion field $\boldsymbol{\phi}$, using the following Jacobian determinant:
\begin{equation}
    det(J_{\boldsymbol{\phi}}(\boldsymbol{p}))=det\left ( \begin{bmatrix}
\frac{\partial \boldsymbol{\phi}_{x}(\boldsymbol{p})}{\partial x} & \frac{\partial \boldsymbol{\phi}_{x}(\boldsymbol{p})}{\partial y} \\ 
\frac{\partial \boldsymbol{\phi}_{y}(\boldsymbol{p})}{\partial x} & \frac{\partial \boldsymbol{\phi}_{y}(\boldsymbol{p})}{\partial y} \\ 
\end{bmatrix} \right ),
\end{equation}
where $\boldsymbol{p}$ is a certain position. Such a Jacobian determinant could be used to analyze the local behavior of the motion field. A positive Jacobian determinant $det(J_{\boldsymbol{\phi}}(\boldsymbol{p}))$ indicates the motion field at position $\boldsymbol{p}$ preserves the orientation in the neighborhood of $\boldsymbol{p}$. However, a negative Jacobian determinant $det(J_{\boldsymbol{\phi}}(\boldsymbol{p}))$ indicates the motion field at position $\boldsymbol{p}$ reverses the orientation in the neighborhood of $\boldsymbol{p}$, which will lose the one-to-one mapping.

\subsection{More Detailed Results}
\subsubsection{T-MRI Image Sequence Registration Results}
In the supplementary folder \verb'Registration', we show representative t-MRI image sequence registration results: (upper-left) tagging image sequence; (upper-right) forward registration results; (bottom-left) backward registration results; (bottom-right) Lagrangian registration results.

\subsubsection{Landmarks Tracking Results}
In the supplementary folder \verb'LM_Tracking', we show representative landmarks tracking results on basal, middle and apical slice: red is ground truth, green is prediction. Note that in the basal slice, on the septum wall, which is between RV and LV, tags may apparently disappear for some frames, due to through-plane motion, as do the ground truth landmarks, but we still showed the predicted landmarks on the closest position. Our method can even track the motion on the last several frames accurately, in spite of the significant image quality degradation.

\subsubsection{Motion Field Results}
In supplementary folder \verb'Motion_Field_Quiver', we show representative INF motion fields and Lagrangian motion fields, represented as a ``quiver" form. Note that our method accurately captures the back-and-forth motion in the left ventricle myocardium wall during systole. Also note that our method can even track the right ventricle's motion accurately.

In the folder \verb'Motion_Field_Map', we show corresponding Lagrangian motion fields: (left) $x$ component; (right) $y$ component.

\subsubsection{Virtual Tag Grid Tracking Results}
In the supplementary folder \verb'Virtual_Tag_Grid', we show representative virtual tag grid tracking results on the short-axis view: (left) tagging image sequence; (middle) warped virtual tag grid by the Lagrangian motion field; (right) virtual tag grid superimposed on tagging images. Note that the virtual tag grid has been aligned with the tag pattern at time $t=0$. As time goes on, the virtual tag grid is deformed by the predicted Lagrangian motion field and follows the underlying tag pattern in the images very well.

In the folder \verb'Virtual_Tag_Grid_LAX', we show representative virtual tag grid tracking results on the long axis (2-, 3-, 4-chamber) views: (upper) tagging image sequence; (bottom) virtual tag grid superimposed on tagging images.

\subsubsection{Full Sequence Motion Tracking Results}
\begin{figure}[t]
\begin{center}
\includegraphics[width=0.97\linewidth]{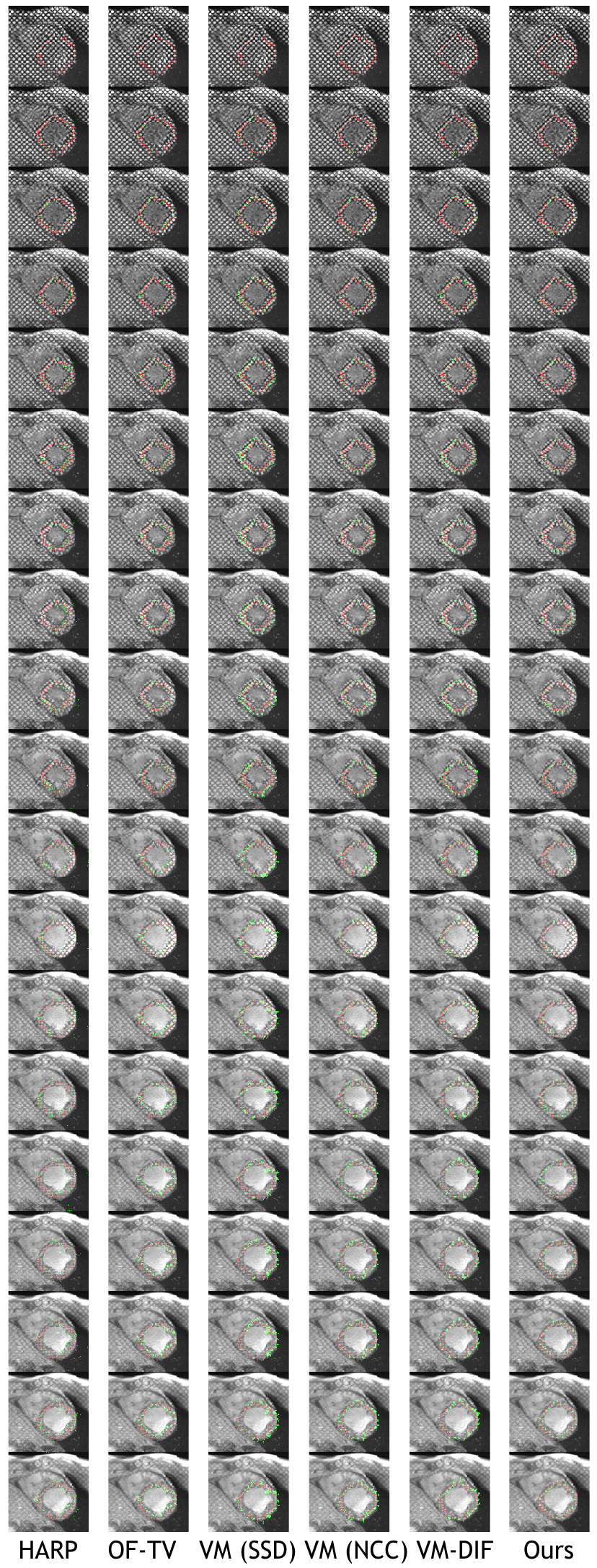}
\end{center}
   \caption{Motion tracking results shown on a full t-MRI image sequence (best viewed zoomed in). Red is ground truth, green is prediction.}
\label{sfig7}
\end{figure}

\begin{figure}[t]
\begin{center}
\includegraphics[width=1.0\linewidth]{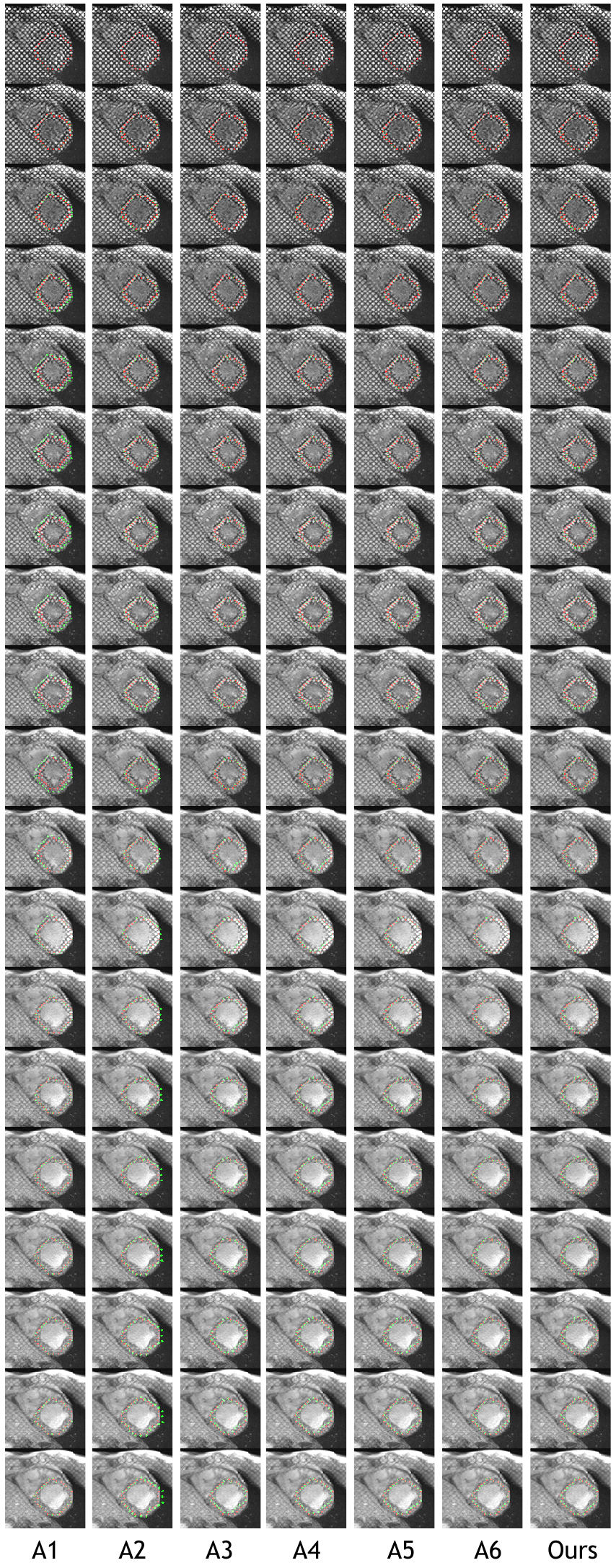}
\end{center}
   \caption{Ablation study motion tracking results shown on a full t-MRI image sequence (best viewed zoomed in). Red is ground truth, green is prediction.}
\label{sfig9}
\end{figure}

In Fig.~\ref{sfig7} and Fig.~\ref{sfig9}, we show the motion tracking results on a full t-MRI image sequence. 

\end{document}